\documentclass[graybox]{svmult}

\usepackage[utf8]{inputenc}                
\usepackage{mathrsfs}                                              
\usepackage[T1]{fontenc}
\usepackage{listings}
\usepackage{color}

\usepackage{bm}
\usepackage{url}
\usepackage{cite}
\usepackage{epsfig}

\usepackage{mathtools}
\usepackage{subfigure}
\usepackage{makecell}
\usepackage{multirow}

\DeclareGraphicsExtensions{.pdf,.png,.jpg,.fig}

\usepackage[cmintegrals]{newtxmath}
\usepackage[width=122mm,left=12mm,paperwidth=146mm,height=193mm,top=12mm,paperheight=217mm]{geometry}
\usepackage[colorlinks=true,citecolor=red,linkcolor=blue]{hyperref}

\makeatletter
\newcommand\footnoteref[1]{\protected@xdef\@thefnmark{\ref{#1}}\@footnotemark}
\makeatother

\newcommand{\etc}{\textit{etc}}
\makeindex 
\begin{document}
	
	\title*{Computer Stereo Vision for Autonomous Driving}
	\author{Rui Fan, Li Wang, Mohammud Junaid Bocus, Ioannis Pitas}
	\institute{Rui Fan \at UC San Diego, \email{rfan@ucsd.edu}
		\and Li Wang \at ATG Robotics, \email{li.wang@ieee.org}
		\and Mohammud Junaid Bocus \at University of Bristol, \email{junaid.bocus@bristol.ac.uk}
		\and Ioannis Pitas \at Aristotle University of Thessaloniki, \email{pitas@csd.auth.gr}
	}
	
	\maketitle

	\abstract{ 
		As an important component of autonomous systems, autonomous car perception has had a big leap with recent advances in parallel computing architectures. With the use of tiny but full-feature embedded supercomputers, computer stereo vision has been prevalently applied in autonomous cars for depth perception. The two key aspects of computer stereo vision are speed and accuracy. They  are both desirable but conflicting properties, as the algorithms with better disparity accuracy usually have higher computational complexity. Therefore, the main aim of developing a computer stereo vision algorithm for resource-limited hardware is to improve the trade-off between speed and accuracy. 
		In this chapter, we introduce both the hardware and software aspects of computer stereo vision for autonomous car systems. Then, we discuss four autonomous car perception tasks, including 1) visual feature detection, description and matching, 2) 3D information acquisition, 3) object detection/recognition and 4) semantic image segmentation. The principles of computer stereo vision and parallel computing on multi-threading CPU and GPU architectures are then detailed. 
	}

	\clearpage

	\section{Introduction}
	\label{sec.intro}
	
\begin{figure*}[h!]
		\vspace{-1.0em}
		\begin{center}
			\centering
			\includegraphics[width=0.96\textwidth]{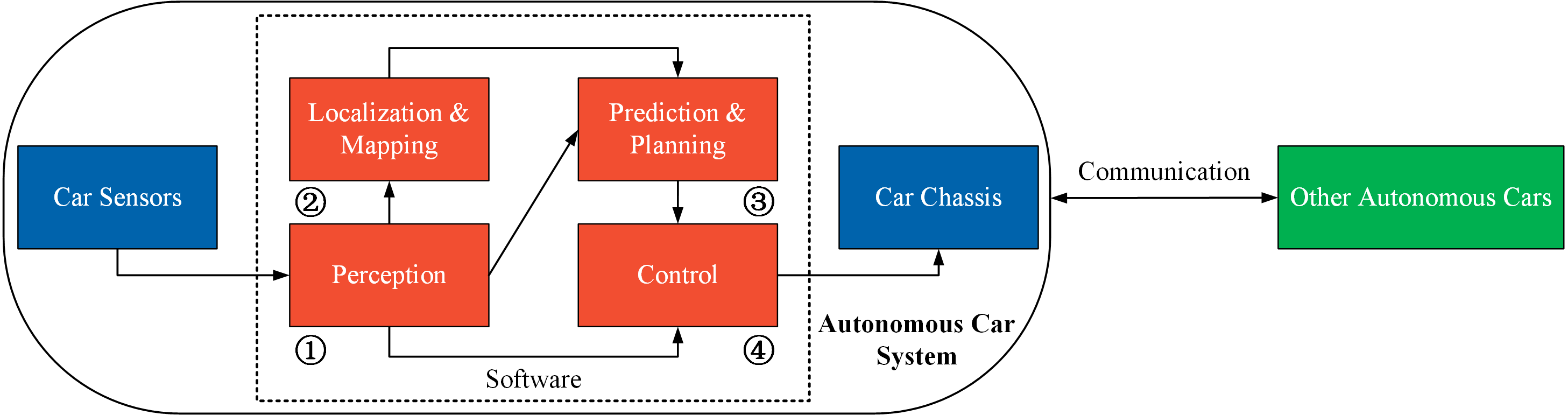}
			\caption{Autonomous car system architecture.
			}
		\end{center}
		\label{fig.autonomous_car_system}
		\vspace{-2.1em}
	\end{figure*}

	Autonomous car systems enable self-driving cars to navigate in complicated environments, without any intervention of human drivers. 
	An example of  autonomous car system architecture is illustrated in Fig \ref{fig.autonomous_car_system}. Its hardware (HW) mainly includes: 1) car sensors, such as cameras, LIDARs, and Radars; and 2) car chassis, such as throttle, brake, and wheel. On the other hand, the software (SW) is comprised of four main functional modules \cite{fan2019key}: 1) perception, 2) localization and mapping, 3) prediction and planning and 4) control. Computer stereo vision is an important component of the perception module. It enables self-driving cars to perceive environment in 3D.

	This chapter first introduces the HW/SW aspects of an autonomous car system. Then, four autonomous car perception tasks are discussed, including: 1) visual feature detection, description and matching, 2) 3D information acquisition, 3) object detection/recognition and 4) semantic image segmentation. 
	Finally, the principles of computer stereo vision and parallel computing on multi-threading CPU and GPU are detailed. 
	
	\section{Autonomous Car System}
	\label{sec.autonomous_systems}
	\subsection{Hardware}
	\subsubsection{Car Sensors}
	\label{sec.sensors}
	The most commonly used car sensors include: a) passive sensors, such as cameras; b) active sensors, such as LIDARs, Radars and ultrasonic transceivers; and c) other types of sensors, such as global positioning systems (GPS), inertial measurement unit (IMU), among others. 
	When choosing them, we need to consider many different factors, such as sampling rate, field of view (FoV), accuracy, range, cost and overall system complexity\footnote{\url{autonomous-driving.org/2019/01/25/positioning-sensors-for-autonomous-vehicles}\label{fn.sensors}}.
	
	Cameras capture 2D images, by collecting light reflected on 3D objects. Images captured from different views can be utilized to reconstruct the 3D driving scene geometry. 
	Most autonomous car perception tasks, such as visual semantic driving scene segmentation and object detection/recognition, are developed for images. In Sec. \ref{sec.autonomous_perception}, we provide readers with a comprehensive overview of these tasks. The perspective (or pinhole) camera model and the mathematical principles of multi-view geometry are discussed in Sec. \ref{sec.computer_stereo_vision}. 
	Acquired image quality is always subject to environmental conditions, such as weather and illumination \cite{pitas2000digital}. Therefore, the visual information fusion from other sensors  is typically required for robust autonomous car perception. 
	
	LIDAR illuminates a target with pulsed laser light and measures the source-target distance, by analyzing the reflected laser pulses \cite{detection2013ranging}. Due to its ability to generate highly accurate 3D driving scene geometry models, LIDARs are generally mounted on autonomous cars for depth perception. Current industrial autonomous car localization and mapping systems are generally based on LIDARs. Furthermore, Radars can measure both the range and radial velocity of an object, by transmitting an electromagnetic wave and analyzing its reflections \cite{bureau2013radar}. Radars have already been established in the automotive industry, and they have been prevalently employed to enable intelligent vehicle  advanced driver assistance system (ADAS) features, such as adaptive cruise control and autonomous emergency braking\footnoteref{fn.sensors}. Similar to Radar, ultrasonic transceivers calculate the source-object distance, by measuring the time between transmitting an ultrasonic signal and receiving its echo \cite{westerveld2014silicon}. Ultrasonic transceivers are commonly used for autonomous car localization and navigation. 
	
	In addition to the aforementioned passive and active sensors, GPS and IMU systems are commonly used to enhance autonomous car localization and mapping performance \cite{zheng2019low}. GPS can provide both time and geolocation information for autonomous cars. However, its signals can become very weak, when GPS reception is blocked by obstacles in GPS-denied regions, such as urban regions \cite{samama2008global}. Hence, GPS and IMU information fusion is widely adopted to provide continuous autonomous car position and velocity information \cite{zheng2019low}. 
	
	\subsubsection{Car Chassis}
	\label{sec.controllers}
	Car chassis technologies, especially Drive-by-Wire (DbW), are required for building autonomous vehicles.  DbW technology refers to the electronic systems that can replace traditional mechanical controls \cite{liu2019chassis}. DbW systems can perform vehicle functions, which are traditionally achieved by mechanical linkages, through electrical or electro-mechanical systems. There are three main vehicle control systems that are commonly replaced with electronic controls: 1) throttle, 2) braking and 3) steering. 
	
	A Throttle-by-Wire (TbW) system helps accomplish vehicle propulsion via an electronic throttle, without any cables from the accelerator pedal to the engine throttle valve. In electric vehicles, TbW system controls the electric motors, by sensing accelerator pedal for a pressure (input) and sending signal to the power inverter modules. Compared to traditional hydraulic brakes, which provide braking effort, by building hydraulic pressure in the brake lines, a Brake-by-Wire (BbW) system  completely eliminates the need for hydraulics,  by using electronic motors to activate calipers. Furthermore, in vehicles that are equipped with Steer-by-Wire (SbW) technology, there is no physical connection between the steering wheel and the tires. The control of wheels' direction is established through electric motor(s), which are actuated by electronic control units monitoring steering wheel inputs.
	
	In comparison to traditional throttle systems, electronic throttle systems are much lighter, hence greatly reducing  modern car weight. In addition, they are  easier to service and tune, as a technician can simply connect a laptop to perform  tuning. Moreover, an electronic control system allows more accurate control of  throttle opening, compared to a cable control that stretches over time. Furthermore, since the steering wheel can be bypassed as an input device, safety can be improved by providing computer controlled intervention of vehicle controls with systems, such as Adaptive Cruise Control and  Electronic Stability Control. 
	
	\subsection{Software}
	Autonomous car perception module analyzes the raw data collected by car sensors (see Sec. \ref{sec.sensors}) and outputs its understanding to the environment. This process is similar to human visual cognition. We discuss different autonomous car perception tasks in Sec. \ref{sec.autonomous_perception}. 
	
	Perception module outputs are then used by other modules. The localization and mapping module not only estimates autonomous car location, but also constructs and updates the 3D environment map \cite{bhuttaloop}. This topic has become very popular, since the concept of \textit{Simultaneous Localization and Mapping (SLAM)} was introduced in 1986 \cite{smith1986representation}. 
	
	Prediction and planning module first analyzes the motion patterns of other traffic agents and predicts their future trajectories. Such prediction outputs are then used to determine possible safe autonomous car navigation routes \cite{katrakazas2015real} using different path planning techniques, such as Dijkstra \cite{cormen2001section}, A-star (or simply A*) \cite{delling2009engineering}, \etc. 
	
	Finally, autonomous car control module sends appropriate commands to car controllers (see Sec. \ref{sec.controllers}), based on its predicted trajectory and the estimated car state. This enables the autonomous car to follow the planned trajectory, as closely as possible. Traditional controllers, such as proportional-integral-derivative (PID) \cite{willis1999proportional}, linear-quadratic regulator (LQR) \cite{goodwin2001control} and model predictive control (MPC) \cite{garcia1989model} are still the most commonly used ones in autonomous car control module.

	\section{Autonomous Car Perception}
	\label{sec.autonomous_perception}
	The autonomous car perception module has four main functionalities: 
	\begin{enumerate}
		\item visual feature detection, description and matching;
		\item 3D information acquisition;
		\item objection detection/recognition; 
		\item semantic image segmentation. 
	\end{enumerate}
	
	Visual feature detectors and descriptors have become very popular research topics in the computer vision and robotics communities. They have been applied in many application domains \cite{hassaballah2016image}, such as image classification \cite{liu2012discriminative}, 3D scene reconstruction \cite{moreels2007evaluation}, object recognition \cite{ dollar2011pedestrian} and visual tracking \cite{danelljan2016discriminative}. The matched visual feature correspondences between two (or more) images can be utilized to establish image relationships \cite{hassaballah2016image}. The most well-known visual features are scale-invariant feature transform (SIFT) \cite{lowe2004distinctive}, speeded up robust feature (SURF) \cite{bay2006surf}, oriented FAST and rotated BRIEF (ORB) \cite{rublee2011orb}, binary robust invariant scalable keypoints (BRISK) \cite{leutenegger2011brisk}, and so forth. 
	
	\begin{figure}[h!]
		\vspace{-1.0em}
		\centering
		\includegraphics[width=0.95\textwidth]{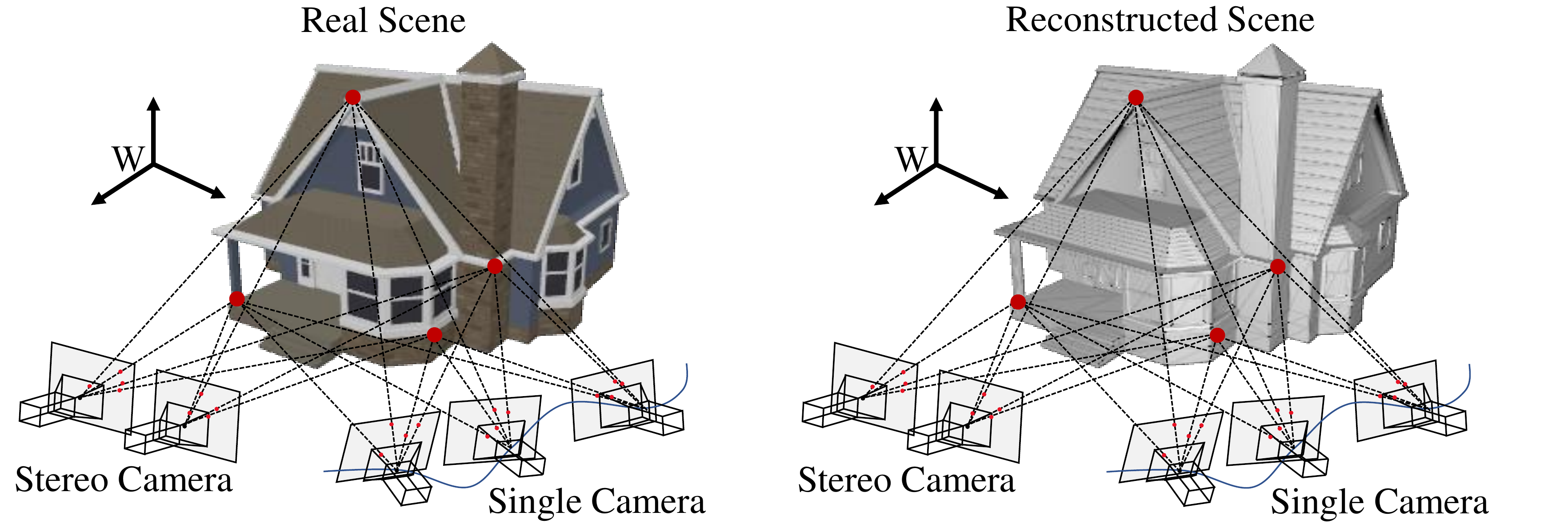}
		\caption{3D scene reconstruction, where $\text{W}$ presents the world coordinate system (WCS).  }
		\label{fig.3d_reconstruction} 
		\vspace{-1.2em}
	\end{figure}

	The digital images captured by cameras are essentially 2D \cite{fan2018real}. In order to extrapolate the 3D information from a given driving scene, images from multiple views are required \cite{hartley2003multiple}. These images can be captured using either a single moving camera \cite{wang2020cot} or an array of synchronized cameras, as shown in Fig. \ref{fig.3d_reconstruction}. The former is typically known as \textit{structure from motion (SfM)} \cite{ullman1979interpretation} or \textit{optical flow} \cite{wang2020cot}, while the latter is typically referred to as \textit{stereo vision} or \textit{binocular vision} (in case two cameras are used) \cite{fan2018real}. SfM methods estimate both camera poses and the 3D points of interest from images captured from multiple views, which are linked by a collection of visual features.  They also leverage bundle adjustment (BA) \cite{triggs1999bundle} technique to refine the estimated camera poses and 3D point locations, by minimizing a cost function known as total re-projection error \cite{wang2020atg}. Optical flow describes the motion of pixels between consecutive frames of a
	video sequence \cite{wang2020atg}. It is also regarded as an effective tool for dynamic object detection \cite{wang2020cot}. Stereo vision acquires depth information by finding the horizontal positional differences (disparities) of the visual feature correspondence pairs between two synchronously captured images. More details on computer stereo vision will be given in Sec. \ref{sec.stereopsis}. 
	\begin{figure}[h!]
		\centering
		\includegraphics[width=0.95\textwidth]{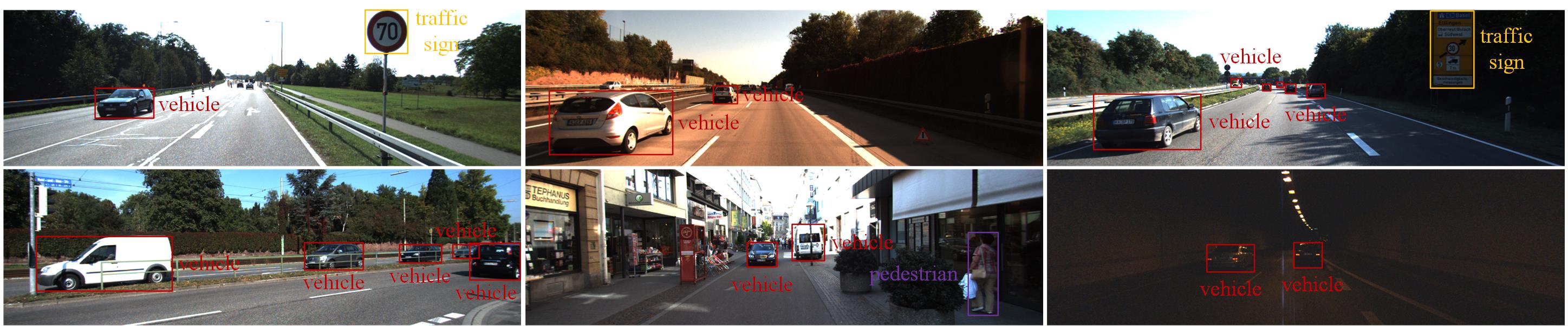}
		\caption{Object detection/recognition.  }
		\vspace{-1.2em}
		\label{fig.object_detection} 
	\end{figure}

	Object detection/recognition refers to the recognition and localization of particular objects in images/videos \cite{zhao2019object}. It can be used in various autonomous car perception subtasks, such as pedestrian detection \cite{wang2019deep}, vehicle detection \cite{wang2020atg}, traffic sign detection \cite{mogelmose2012vision}, cyclist detection \cite{wu2017squeezedet}, \textit{etc}., as shown in Fig. \ref{fig.object_detection}. Object detection approaches can be classified as either computer vision-based or machine/deep learning-based. The former typically consists of three steps \cite{zhao2019object}: 1) informative region selection (scanning the whole image by sliding windows with particular templates to produce candidate regions); 2) visual feature extraction, as discussed above; and 3) object classification (distinguishing a target object from all the other categories using a classifier). With recent advances in machine/deep learning, a large number of convolutional neural networks (CNNs) have been proposed to recognize objects from images/videos. Such CNN-based approaches have achieved very impressive results. The most popular ones include: regions with CNN
	features (R-CNN) \cite{girshick2014rich}, fast R-CNN \cite{girshick2015fast}, faster R-CNN \cite{ren2015faster}, you only look once (YOLO) \cite{redmon2016you}, YOLOv3 \cite{redmon2018yolov3}, YOLOv4 \cite{bochkovskiy2020yolov4}, \textit{etc}. 
	
	\begin{figure}[h!]
		\vspace{-1.0em}
		\centering
		\includegraphics[width=0.95\textwidth]{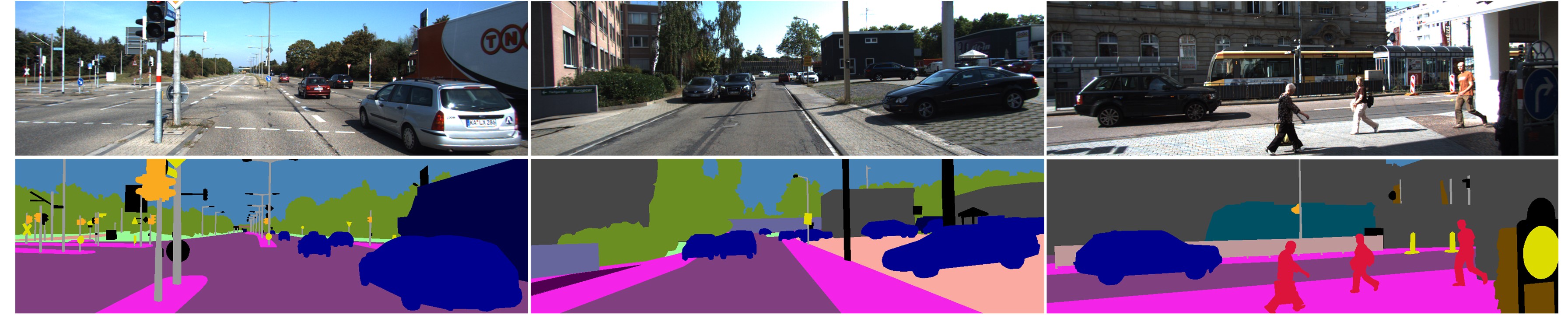}
		\caption{Semantic image segmentation.  }
		\vspace{-1.2em}
		\label{fig.semantic_seg} 
	\end{figure}
	Semantic image segmentation labels every pixel in the image with a given object class \cite{fan2020sne}, such as lane marking, vehicle, collision-free space, or pedestrian, as illustrated in Fig. \ref{fig.semantic_seg}. The state-of-the-art semantic image segmentation approaches are mainly categorized into two main groups \cite{wang2020applying}: 1) single-modal and 2) data-fusion. The former typically segments RGB images with an encoder-decoder CNN architecture \cite{fan2020learning}. In recent years, many popular single-model semantic image segmentation algorithms, such as 
	Fully Convolutional Network (FCN) \cite{long2015fully}, U-Net \cite{ronneberger2015u}, SegNet \cite{badrinarayanan2017segnet}, DeepLabv3+ \cite{chen2018encoder}, DenseASPP \cite{yang2018denseaspp}, DUpsampling \cite{tian2019decoders}, \textit{etc.}, have been proposed. Data-fusion semantic image segmentation approaches  generally learn features from two different types of vision data \cite{fan2020we}, such as RGB and depth images in FuseNet \cite{hazirbas2016fusenet}, RGB and surface normal images \cite{fan2020three} in SNE-RoadSeg \cite{fan2020sne}, RGB and transformed disparity \cite{fan2018ipl, fan2019pothole, fan2019road} images in AA-RTFNet \cite{fan2020we}, or RGB and thermal images in MFNet\cite{ha2017mfnet}. The learned feature maps are then fused to provide a better semantic prediction. 
	
	Please note: a given autonomous car perception application can always be solved by different types of techniques. For instance, lane marking detection can be formulated as a linear/quadratic/quadruplicate pattern recognition problem and solved using straight line detection algorithms \cite{fan2016ist} or dynamic programming algorithms \cite{ozgunalp2016multiple, fan2018lane, jiao2019using}. On the other hand, it can also be formulated as a semantic image segmentation problem and solved with CNNs.

	\section{Computer Stereo Vision}
	\label{sec.computer_stereo_vision}
	\subsection{Preliminaries}
	\label{sec.preliminaries}
	\subparagraph{{1. Skew-symmetric matrix}}
	\label{sec.skew_symmetric}
	In linear algebra, a \textit{skew-symmetric matrix} $\mathbf{A}$ satisfies the following property:  its transpose is identical to its negative, \textit{i.e.}, $\mathbf{A}^\top=-\mathbf{A}$. In 3D computer vision, the skew-symmetric matrix $[\mathbf{a}]_{\times}$ of a vector $\mathbf{a}=[a_1,a_2,a_3]^\top$ can be written as \cite{hartley2003multiple}:
	\begin{equation}
		[\mathbf{a}]_{\times}=\begin{bmatrix*}[c]
			\ 0 & -a_3 & \ \ a_2\\
			\ \ a_3 & 0 & -a_1\\
			-a_2 & \ \ a_1 & \ \ 0\\
		\end{bmatrix*}.
	\end{equation}
	A skew-symmetric matrix has two important properties:
	\begin{equation}
		\mathbf{a}^\top[\mathbf{a}]_{\times}=\textbf{0}^\top, \ \ \  [\mathbf{a}]_{\times}\mathbf{a}=\textbf{0},
		\label{eq.skew_symmetrix_prop}
	\end{equation}
	where $\textbf{0}=[0,0,0]^\top$ is a zero vector. Furthermore, the cross product of two vectors $\mathbf{a}$ and $\mathbf{b}$ can be formulated as a matrix multiplication process \cite{hartley2003multiple}: 
	\begin{equation}
		\mathbf{a}\times\mathbf{b}=[\mathbf{a}]_{\times}\mathbf{b}=-[\mathbf{b}]_{\times}\mathbf{a}.
		\label{eq.skew_symmetric}
	\end{equation}
	These properties are generally used to simplify the equations related to vector cross-product.

	\subparagraph{2. Lie group SO(3) and SE(3)}
	\label{lie_group}
	A 3D point $\mathbf{x_1}=[x_1,y_1,z_1]^\top\in\mathbb{R}^{3\times1}$ can be transformed into another 3D point $\mathbf{x_2}=[x_2,y_2,z_2]^\top\in\mathbb{R}^{3\times1}$ using a rotation matrix $\mathbf{R}\in\mathbb{R}^{3\times3}$ and a translation vector $\mathbf{t}\in\mathbb{R}^{3\times1}$: 
	\begin{equation}
		\mathbf{x_2}=\mathbf{R}\mathbf{x_1}+\mathbf{t}.
		\label{eq.basic_transform}
	\end{equation}
	$\mathbf{R}$ satisfies matrix orthogonality:
	\begin{equation}
		\mathbf{R}\mathbf{R}^\top=\mathbf{R}^\top\mathbf{R}=\mathbf{I}\ \ \text{and} \ \  |\text{det}(\mathbf{R})|=1,
		\label{eq.so_3_conditions}
	\end{equation}
	where $\mathbf{I}$ is an identity matrix and $\text{det}(\mathbf{R})$ represents the determinant of $\mathbf{R}$. The group containing all rotation matrices is referred to as a \textit{special orthogonal group} and is denoted as SO(3). $\mathbf{\tilde{x}_1}=[\mathbf{x_1}^\top,1]^\top$ and $\mathbf{\tilde{x}_2}=[\mathbf{x_2}^\top,1]^\top$, the homogeneous coordinates of $\mathbf{x_1}$ and $\mathbf{x_2}$,  can be used to describe rotation and translation, as follows:
	\begin{equation}
		\mathbf{\tilde{x}_2}=\mathbf{P}\mathbf{\tilde{x}_1},
		\label{eq.basic_transform1}
	\end{equation}
	where
	\begin{equation}
		\mathbf{P}=
		\left[
		\begin{array}{c|c}
			\mathbf{R} & \mathbf{t} \\
			\hline
			\textbf{0}^\top  & 1
		\end{array}\right],
	\end{equation}
	$\mathbf{P}$ is a homogeneous transformation matrix\footnote{\url{seas.upenn.edu/~meam620/slides/kinematicsI.pdf}}. The group containing all homogeneous transformation matrices is referred to as a \textit{special Euclidean group} and is denoted as SE(3).
	\subsection{Multi-View Geometry}
	\label{sec.multi_view_geometry}
	\subsubsection{Perspective Camera Model}
	
\begin{figure}[h!]
		\vspace{-1.0em}
		\centering
		\includegraphics[width=0.400\textwidth]{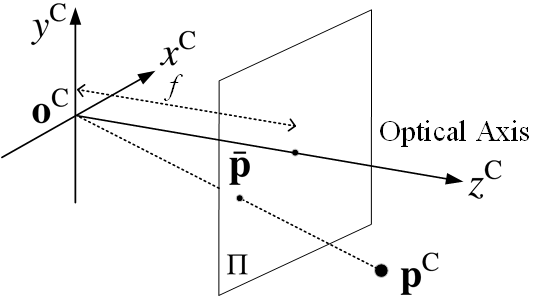}
		\caption{Perspective camera model. }
		\label{fig.perspective_camera} 
		\vspace{-1.0em}
	\end{figure}

	The perspective (or pinhole) camera model, as illustrated in Fig. \ref{fig.perspective_camera}, is the most common  geometric camera model describing the relationship between a 3D point $\mathbf{p^\text{C}}=[x^\text{C},y^\text{C},z^\text{C}]^\top$ in the camera coordinate system {(CCS)} and its projection $\mathbf{\bar{p}}=[x,y,f]^\top$ on the image plane $\Pi$. $\mathbf{o^\text{C}}$ is the camera center. The distance between $\Pi$ and $\mathbf{o^\text{C}}$ is the camera focal length $f$.
	$\hat{\mathbf{p}}^\text{C}=[\frac{x^\text{C}}{z^\text{C}},\frac{y^\text{C}}{z^\text{C}},1]^\top$ are the normalized coordinates of $\mathbf{p^\text{C}}=[x^\text{C},y^\text{C},z^\text{C}]^\top$.  Optical axis is the ray originating from $\mathbf{o^\text{C}}$ and passing perpendicularly through $\Pi$.  
	The relationship between $\mathbf{p}^\text{C}$ and $\bar{\mathbf{p}}$ is as follows \cite{trucco1998introductory}:
	\begin{equation}
		\mathbf{\bar{p}}=f\hat{\mathbf{p}}^\text{C}=\frac{f}{z^\text{C}}\mathbf{p^\text{C}}.
		\label{eq.perspective_camera}
	\end{equation}
	\subsubsection{Intrinsic Matrix}
	Since lens distortion does not exist in a perspective camera model, $\mathbf{\bar{p}}=[x,y,f]^\top$ on the image plane $\Pi$ can be transformed into a pixel $\mathbf{{p}}=[u,v]^\top$ in the image using \cite{fan2018real}:
	\begin{equation}
		u=u_o+s_x x, \ \ \
		v=v_o+s_y y,
		\label{eq.ccs2ics}
	\end{equation}
	where $\mathbf{p}_o=[u_o,v_o]^\top$ is the principal point; and $s_x$ and $s_y$ are the effective size measured (in pixels per millimeter) in the horizontal and vertical directions, respectively \cite{trucco1998introductory}. To simplify the expression of the camera intrinsic matrix $\mathbf{K}$, two notations $f_x=fs_x$ and $f_y=fs_y$ are introduced. $u_o$, $v_o$, $f$, $s_x$ and $s_y$ \cite{hartley2003multiple} are five camera intrinsic parameters.  Combining (\ref{eq.perspective_camera}) and 
	(\ref{eq.ccs2ics}), a 3D point $\mathbf{p}^\text{C}$ in the CCS can be transformed into a pixel $\mathbf{p}$ in the image using \cite{fan2019cvprw}:
	\begin{equation}
		\begin{split}
			\mathbf{\tilde{p}}=\frac{1}{z^\text{C}}\mathbf{K}\mathbf{p^\text{C}}=\frac{1}{z^\text{C}}\begin{bmatrix}
				f_x & 0 & u_o\\
				0 & f_y & v_o\\
				0 & 0 & 1\\
			\end{bmatrix}\begin{bmatrix}
				x^\text{C}\\
				y^\text{C}\\
				z^\text{C}\\
			\end{bmatrix}
		\end{split},
		\label{eq.K}
	\end{equation}
	where $\tilde{\mathbf{p}}=[\mathbf{{p}}^\top, 1]^\top=[u,v,1]^\top$ denotes the homogeneous coordinates of $\mathbf{p}=[u,v]^\top$. Plugging (\ref{eq.K}) into (\ref{eq.perspective_camera}) results in:
	\begin{equation}
		\hat{\mathbf{p}}^\text{C}=\mathbf{K}^{-1}\mathbf{\tilde{p}}=\frac{\mathbf{\bar{p}}}{f}=\frac{\mathbf{p^\text{C}}}{z^\text{C}}.
		\label{eq.normalised_image_point}
	\end{equation}
	Therefore, an arbitrary 3D point lying on the ray, which goes from $\mathbf{o}^\text{C}$ and through $\mathbf{p}^\text{C}$, is always projected at $\bar{\mathbf{p}}$ on the image plane. 
	\subsubsection{Lens Distortion}
	
\begin{figure}[h!]
		\centering
		\includegraphics[width=0.60\textwidth]{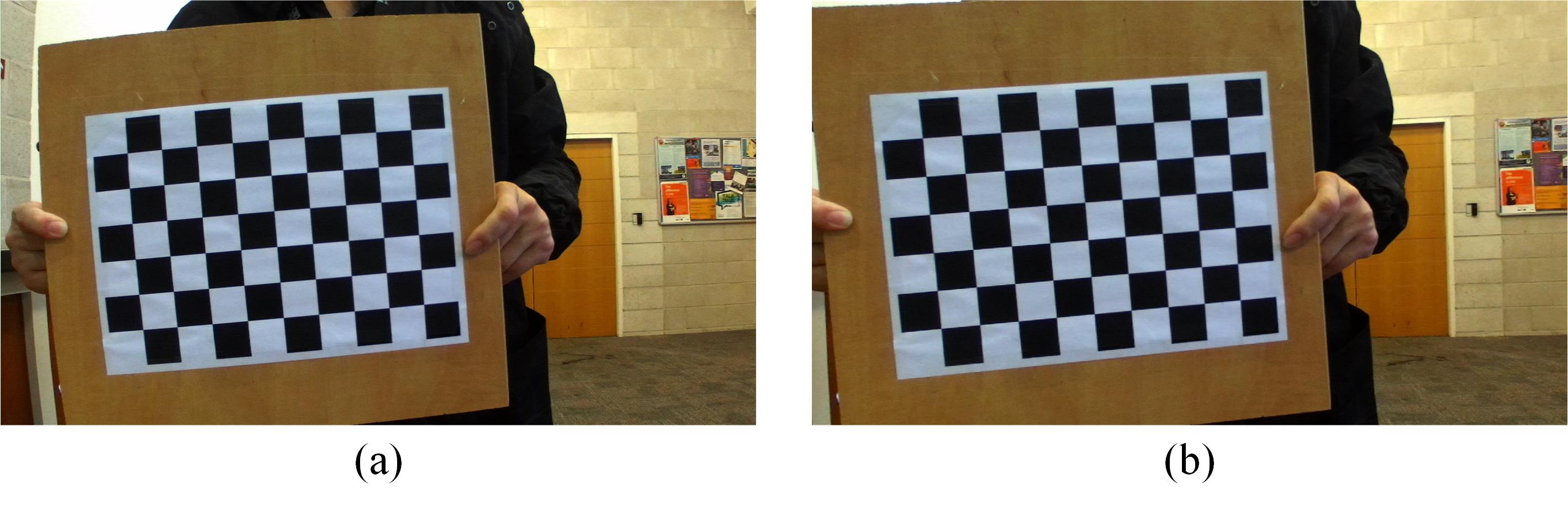}
		\caption{Distorted image correction: (a) original image; (b) corrected image. }
		\label{fig.checkerboard} 
		\vspace{-1.2em}
	\end{figure}

	In order to get better imaging results, a lens is usually installed in front of the camera \cite{trucco1998introductory}. However, this introduces image distortions. The optical aberration caused by the installed lens typically deforms the physically straight lines provided by projective geometry to curves in the images \cite{zhang2000flexible}, as shown in Fig. \ref{fig.checkerboard}(a). We can observe in Fig. \ref{fig.checkerboard}(b) that the bent checkerboard grids become straight when the lens distortion is corrected. 
	
\begin{figure}[h!]
		\vspace{-1.5em}
		\centering
		\includegraphics[width=0.80\textwidth]{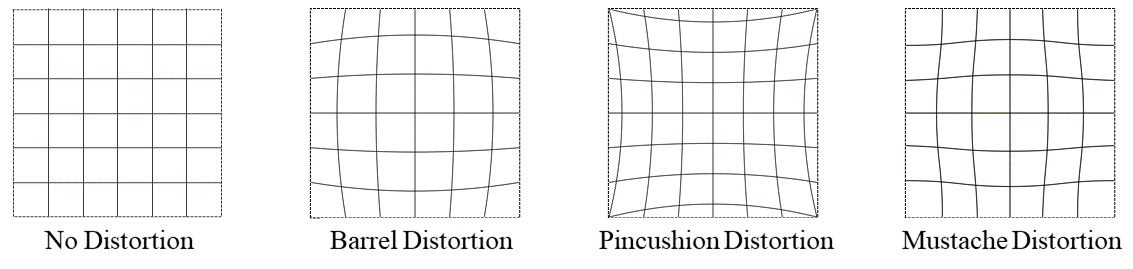}
		\caption{Radial distortion types.}
		\label{fig.lens_distortion} 
		\vspace{-1.5em}
	\end{figure}

	Lens distortion can be categorized into two main types: 1) radial distortion and 2) tangential distortion \cite{fan2018real}. The presence of radial distortion is due to the fact that geometric lens shape affects straight lines. Tangential distortion occurs because the lens is not perfectly parallel to the image plane \cite{trucco1998introductory}. In practical experiments, the image geometry is affected by radial distortion to a much higher extent than by tangential distortion. Therefore, the latter is sometimes neglected in the process of distorted image correction. 
	\paragraph{\textbf{Radial distortion}}
	\label{sec.radial_distortion}
	Radial distortion mainly includes 1) barrel distortion, 2) pincushion distortion and 3) mustache distortion, as illustrated in Fig. \ref{fig.lens_distortion}.
	It can be observed that a) radial distortions are symmetric about the image center and b) straight lines are no longer preserved. 
	In barrel distortion, the image magnification decreases with the distance from the optical axis (lines curve outwards). In contrast, the pincushion distortion pinches the image (lines curve inwards). Mustache distortion is a mixture of the above two distortion types. It starts out as barrel distortion close to the optical axis and gradually turns into pincushion distortion close to image periphery. Barrel distortion is commonly applied in fish-eye lenses to produce wide-angle/panoramic images, while pincushion distortion is often associated with telephoto/zoom lenses. Radial distortions can be corrected using\footnote{\url{docs.opencv.org/2.4/doc/tutorials/calib3d/camera_calibration/camera_calibration.html}}:
	\begin{equation}
		\begin{split}
			x_\text{undist}&=x_\text{dist}(1+k_1r^2+k_2r^4+k_3r^6),\\
			y_\text{undist}&=y_\text{dist}(1+k_1r^2+k_2r^4+k_3r^6),
		\end{split}
		\label{eq.radial_distortion}
	\end{equation}
	where the corrected  point will be  $\mathbf{p}_\text{undist}=[x_\text{undist}, y_\text{undist}]^\top$; $r^2={x_\text{dist}}^2+{y_\text{dist}}^2$; $x_\text{dist}=\frac{x^\text{C}}{z^\text{C}}=\frac{u-u_o}{f_x}$ and  $y_\text{dist}=\frac{y^\text{C}}{z^\text{C}}=\frac{v-v_o}{f_y}$\footnote{\url{docs.opencv.org/2.4/modules/imgproc/doc/geometric_transformations.html}} can be obtained from the distorted image.
	$k_1$, $k_2$ and $k_3$ are three intrinsic parameters used for radial distortion correction. They can be estimated using a collection of images containing a planar checkerboard pattern.  
	\paragraph{\textbf{Tangential distortion}}
	\label{sec.tangential_distortion}
	Similar to radial distortion, tangential distortion can also be corrected using:
	\begin{equation}
		\begin{split}
			x_\text{undist}&=x_\text{dist}+\big[2p_1x_\text{dist}y_\text{dist}+p_2(r^2+2{x_\text{dist}}^2)\big],\\
			y_\text{undist}&=y_\text{dist}+\big[p_1(r^2+2{y_\text{dist}}^2)+2p_2x_\text{dist}y_\text{dist}\big],
		\end{split}
		\label{eq.tangential_distortion}
	\end{equation}
	where $p_1$ and $p_2$ are two intrinsic parameters, which can also be estimated using a collection of images containing a planar checkerboard pattern. 
	
	\subsubsection{Epipolar Geometry}
	\label{sec.epipolar_geometry}
\begin{figure}[h!]
		\centering
		\includegraphics[width=0.55\textwidth]{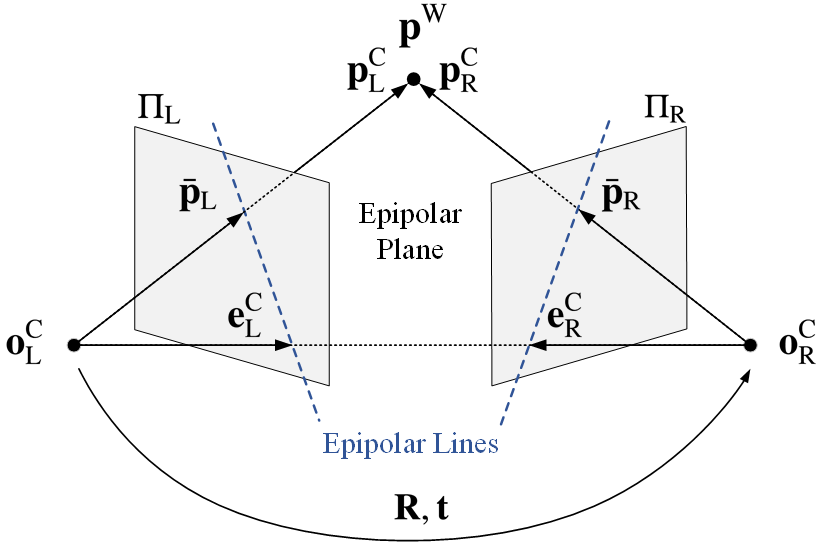}
		\caption{Epipolar geometry.  }
		\vspace{-1.35em}
		\label{fig.epipolar_geometry} 
	\end{figure}

	The generic geometry of stereo vision is known as \textit{epipolar geometry}. An example of the epipolar geometry is shown in Fig. \ref{fig.epipolar_geometry}.
	$\Pi_\text{L}$ and $\Pi_\text{R}$ represent the left and right image planes, respectively. $\mathbf{o}_\text{L}^\text{C}$ and $\mathbf{o}_\text{R}^\text{C}$ denote the origins of the left camera coordinate system (LCCS) and the right camera coordinate system (RCCS), respectively. The 3D point $\mathbf{p}^\text{W}=[x^\text{W},y^\text{W},z^\text{W}]^\top$ in the 
	WCS, is projected at $\mathbf{\bar{p}_\text{L}}=[x_\text{L},y_\text{L},f_\text{L}]^\top$ on $\Pi_\text{L}$ and at $\mathbf{\bar{p}_\text{R}}=[x_\text{R},y_\text{R},f_\text{R}]^\top$ on $\Pi_\text{R}$, respectively. $f_\text{L}$ and $f_\text{R}$ are the focal lengths of the left and right cameras, respectively;  The representations of $\mathbf{p}^\text{W}$ in the LCCS and RCCS are $\mathbf{p^\text{C}_\text{L}}=[x^\text{C}_\text{L},y^\text{C}_\text{L},z^\text{C}_\text{L}]^\top=\frac{{z_\text{L}^\text{C}}}{f_\text{L}}\mathbf{\bar{p}_\text{L}}$ and $\mathbf{p^\text{C}_\text{R}}=[x^\text{C}_\text{R},y^\text{C}_\text{R},z^\text{C}_\text{R}]^\top=\frac{{z_\text{R}^\text{C}}}{f_\text{R}}\mathbf{\bar{p}_\text{R}}$, respectively. 
	According to (\ref{eq.basic_transform}), $\mathbf{p^\text{C}_\text{L}}$ can be transformed into $\mathbf{p^\text{C}_\text{R}}$ using:
	\begin{equation}
		\mathbf{p^\text{C}_\text{R}}=\mathbf{R}\mathbf{p^\text{C}_\text{L}}+\mathbf{t},
		\label{eq.RT}
	\end{equation}
	where $\mathbf{R}\in\mathbb{R}^{3\times 3}$ is a rotation matrix and $\mathbf{t}\in\mathbb{R}^{3\times 1}$ is a translation vector. $\mathbf{e^\text{C}_\text{L}}$ and $\mathbf{e^\text{C}_\text{R}}$ denote the left and right \textit{epipoles}, respectively. The \textit{epipolar plane} is uniquely defined by  $\mathbf{o^\text{C}_\text{L}}$, $\mathbf{o^\text{C}_\text{R}}$ and $\mathbf{p^\text{W}}$. It
	intersects $\Pi_\text{L}$ and $\Pi_\text{R}$ giving rise to two \textit{epipolar lines}, as shown in Fig. \ref{fig.epipolar_geometry}. Using (\ref{eq.normalised_image_point}), $\mathbf{p^\text{C}_\text{L}}$ and $\mathbf{p^\text{C}_\text{R}}$ can be normalized using: 
	\begin{equation}
		\begin{split}
			\mathbf{\hat{p}^\text{C}_\text{L}}=\frac{\mathbf{p^\text{C}_\text{L}}}{z^\text{C}_\text{L}}={\mathbf{K}_\text{L}}^{-1}\mathbf{\tilde{p}}_\text{L},\ \ \ 
			\mathbf{\hat{p}^\text{C}_\text{R}}=\frac{\mathbf{p^\text{C}_\text{R}}}{Z^\text{C}_\text{R}}=\mathbf{K_\text{R}}^{-1}\mathbf{\tilde{p}_\text{R}},
		\end{split}
		\label{eq.left_right_normalised_image_point}
	\end{equation}
	where $\mathbf{K_\text{L}}$ and $\mathbf{K_\text{R}}$ denote the intrinsic matrices of the left and right cameras, respectively. $\mathbf{\tilde{p}}_\text{L}=[{\mathbf{p}_\text{L}}^\top,1]^\top$ and $\mathbf{\tilde{p}}_\text{R}=[{\mathbf{p}_\text{R}}^\top,1]^\top$ are the homogeneous coordinates of the image pixels $\mathbf{p}_\text{L}=[u_\text{L},v_\text{L}]^\top$ and $\mathbf{p}_\text{R}=[u_\text{R},v_\text{R}]^\top$, respectively. 
	\subsubsection{Essential Matrix}
	\label{sec.essential_mat}
	\textit{Essential matrix} $\mathbf{E}\in \mathbb{R}^{3\times 3}$ was first introduced by Longuet-Higgins in 1981  \cite{longuet1981computer}. 
	A simple way of introducing the defining equation of $\mathbf{E}$ is to multiply  both sides of (\ref{eq.RT}) by $\mathbf{p^\text{C}_\text{R}}^\top[\mathbf{t}]_{\times}$: 
	\begin{equation}
		\begin{split}
			\mathbf{p^\text{C}_\text{R}}^\top[\mathbf{t}]_{\times}\mathbf{p^\text{C}_\text{R}}=\mathbf{p^\text{C}_\text{R}}^\top[\mathbf{t}]_{\times}(\mathbf{R}\mathbf{p^\text{C}_\text{L}}+\mathbf{t}).
		\end{split}
		\label{eq.RT1}
	\end{equation}
	According to (\ref{eq.skew_symmetric}), (\ref{eq.RT1}) can be rewritten as follows: 
	\begin{equation}
		-\mathbf{p^\text{C}_\text{R}}^\top[\mathbf{p^\text{C}_\text{R}}]_{\times}\mathbf{t}=\mathbf{p^\text{C}_\text{R}}^\top[\mathbf{t}]_{\times}\mathbf{R}\mathbf{p^\text{C}_\text{L}}+\mathbf{p^\text{C}_\text{R}}^\top[\mathbf{t}]_{\times}\mathbf{t}.
		\label{eq.RT2}
	\end{equation}
	Applying (\ref{eq.skew_symmetrix_prop}) to (\ref{eq.RT2}) yields: 
	\begin{equation}
		\mathbf{p^\text{C}_\text{R}}^\top[\mathbf{t}]_{\times}\mathbf{R} \mathbf{p^\text{C}_\text{L}}=\mathbf{p^\text{C}_\text{R}}^\top\mathbf{E} \mathbf{p^\text{C}_\text{L}}=0,
		\label{eq.essential_mat_deduction}
	\end{equation}
	The essential matrix $\mathbf{E}$ is defined by:
	\begin{equation}
		\mathbf{E}=[\mathbf{t}]_{\times}\mathbf{R}
		\label{eq.E}
	\end{equation}
	Plugging (\ref{eq.left_right_normalised_image_point}) into (\ref{eq.essential_mat_deduction}) results in:
	\begin{equation}
		{\hat{\mathbf{p}}^\text{C}_\text{R}}{}^\top\mathbf{E} {\hat{\mathbf{p}}^\text{C}_\text{L}}=0,
		\label{eq.essential_mat_deduction2}
	\end{equation}
	which depicts the relationship between each pair of normalized points ${\hat{\mathbf{p}}^\text{C}_\text{R}}$ and ${\hat{\mathbf{p}}^\text{C}_\text{L}}$ lying on the same epipolar plane. It is important to note here that $\mathbf{E}$ has five degrees of freedom: both $\mathbf{R}$ and $\mathbf{t}$ have three degrees of freedom, but the overall scale ambiguity causes the degrees of freedom to be reduced by one \cite{hartley2003multiple}. Hence, in theory, $\mathbf{E}$ can be estimated with at least five pairs of $\mathbf{{p^\text{C}_\text{L}}}$ and $\mathbf{{p^\text{C}_\text{R}}}$. However, due to the non-linearity of $\mathbf{E}$, its estimation using five pairs of correspondences is always intractable. Therefore, $\mathbf{E}$ is commonly estimated with at least eight pairs of $\mathbf{{p^\text{C}_\text{L}}}$ and $\mathbf{{p^\text{C}_\text{R}}}$ \cite{trucco1998introductory}, as discussed in Sec. \ref{sec.fundamental_matrix}.
	
	\subsubsection{Fundamental Matrix}
	\label{sec.fundamental_matrix}
	As introduced in Sec. \ref{sec.essential_mat}, the essential matrix creates a link between a given pair of corresponding 3D points in the LCCS and RCCS. When the intrinsic matrices of the two  cameras in a stereo rig are unknown, the relationship between each pair of corresponding 2D image pixels $\mathbf{p}_\text{L}=[u_\text{L},v_\text{L}]^\top$ and $\mathbf{p}_\text{R}=[u_\text{R},v_\text{R}]^\top$ can be established, using the  \textit{fundamental matrix} $\mathbf{F}\in	\mathbb{R}^{3\times 3}$. It can be considered as a generalization of $\mathbf{E}$, where the assumption of calibrated cameras is removed \cite{hartley2003multiple}. Applying  (\ref{eq.left_right_normalised_image_point}) to (\ref{eq.essential_mat_deduction2}) yields:
	\begin{equation}
		{\tilde{\mathbf{p}}_\text{R}}^\top{\mathbf{K}_\text{R}}^{-\top}\mathbf{E} {\mathbf{K}_\text{L}}^{-1}\tilde{\mathbf{p}}_\text{L}={\tilde{\mathbf{p}}_\text{R}}^\top\mathbf{F}\tilde{\mathbf{p}}_\text{L}=0,
		\label{eq.fundamental_mat_deduction}
	\end{equation}
	where the fundamental matrix $\mathbf{F}$ is defined as:
	\begin{equation}
		\mathbf{F}={\mathbf{K}_\text{R}}^{-\top}\mathbf{E} {\mathbf{K}_\text{L}}^{-1}.
	\end{equation}
	$\mathbf{F}$ has seven degrees of freedom: a 3$\times$3 homogeneous matrix has eight independent ratios, as there are nine entries, but the common scaling is not significant. However, $\mathbf{F}$ also satisfies the constraint $\text{det}(\mathbf{F})= 0$, which removes
	one degree of freedom \cite{hartley2003multiple}.
	The most commonly used algorithm to estimate $\mathbf{E}$ and $\mathbf{F}$ is ``eight-point algorithm'' (EPA), which was introduced by Hartley in 1997 \cite{hartley1997defense}. This algorithm is based on the scale invariance of  $\mathbf{E}$ and $\mathbf{F}$:  $\lambda_\mathbf{E}{\mathbf{p}^\text{C}_\text{R}}^\top\mathbf{E}{\mathbf{p}^\text{C}_\text{L}}=0$ and  $\lambda_\mathbf{F}{\tilde{\mathbf{p}}_\text{R}}^\top \mathbf{F} {\tilde{\mathbf{p}}_\text{L}}=0$, where $\lambda_\mathbf{E},\lambda_\mathbf{F}\neq0$. By setting one element in $\mathbf{E}$ and $\mathbf{F}$ to $1$, eight unknown elements still need to be estimated. This can be done using at least eight correspondence pairs. If the intrinsic matrices $\mathbf{K}_\text{L}$ and $\mathbf{K}_\text{R}$ of the two cameras are known, the EPA only needs to be carried out once to estimate either $\mathbf{E}$ or $\mathbf{F}$, because the other one can be easily worked out using (\ref{eq.fundamental_mat_deduction}). 
	\subsubsection{Homography Matrix}
	\label{sec.homograph_mat}
	An arbitrary 3D point $\mathbf{p^\text{W}}=[x^\text{W}, y^\text{W}, z^\text{W}]^\top$ lying on a planar surface satisfies: 
	\begin{equation}
		\mathbf{n}^\top \mathbf{p^\text{W}}+b=0,
	\end{equation}
	where $\mathbf{n}=[n_x,n_y,n_z]^\top$ is the normal vector of the planar surface. Its corresponding pixels $\mathbf{p}_\text{L}=[u_\text{L},v_\text{L}]^\top$ and $\mathbf{p}_\text{R}=[u_\text{R},v_\text{R}]^\top$ in the left and right images, respectively, can be linked by a \textit{homography matrix} $\mathbf{H}\in \mathbb{R}^{3\times 3}$. The expression of the planar surface can be rearranged as follows:
	\begin{equation}
		-{\mathbf{n}^\top\mathbf{p^\text{W}}}/{b}=1.
		\label{eq.planar_surface}
	\end{equation}
	Assuming that $\mathbf{p}^\text{C}_\text{L}=\mathbf{p}^\text{W}$ and plugging (\ref{eq.planar_surface}) and (\ref{eq.left_right_normalised_image_point}) into (\ref{eq.RT}) results in:
	\begin{equation}
		\begin{split}
			\mathbf{{p^\text{C}_\text{R}}}=\mathbf{R}\mathbf{{p^\text{C}_\text{L}}}-\frac{1}{b}\mathbf{t}{\mathbf{n}^\top\mathbf{p^\text{C}_\text{L}}}=\bigg(\mathbf{R}-\frac{1}{b}\mathbf{t}{\mathbf{n}^\top}\bigg){z^\text{C}_\text{L}}{\mathbf{K}_\text{L}}^{-1}\tilde{\mathbf{p}}_\text{L}= {z^\text{C}_\text{R}}{\mathbf{K}_\text{R}}^{-1}\tilde{\mathbf{p}}_\text{R}
		\end{split}
		\label{eq.RT3}
	\end{equation}
	Therefore, 
	$\tilde{\mathbf{p}}_\text{L}$ and $\tilde{\mathbf{p}}_\text{R}$ can be linked using:
	\begin{equation}
		\tilde{\mathbf{p}}_\text{R}=\frac{z^\text{C}_\text{L}}{z^\text{C}_\text{R}}{\mathbf{K}_\text{R}}\bigg(\mathbf{R}-\frac{1}{b}\mathbf{t}{\mathbf{n}^\top}\bigg){\mathbf{K}_\text{L}}^{-1}\tilde{\mathbf{p}}_\text{L}=\mathbf{H}\tilde{\mathbf{p}}_\text{L}.
		\label{eq.homograph_mat}
	\end{equation} 
	The homography matrix $\mathbf{H}$ is generally used to distinguish obstacles from a planar surface \cite{fan2018road}. For a well-calibrated stereo vision system, $\mathbf{R}$, $\mathbf{t}$, $\mathbf{K}_\text{L}$ as well as $\mathbf{K}_\text{R}$ are already known, and $z^\text{C}_\text{L}$ is typically equal to $z^\text{C}_\text{R}$. Thus, $\mathbf{H}$ only relates to $\mathbf{n}$ and $b$, and it can be estimated with at least four pairs of correspondences $\mathbf{p}_\text{L}$ and $\mathbf{p}_\text{R}$ \cite{fan2018road}.
	
	\subsection{Stereopsis}
	\label{sec.stereopsis}
	\subsubsection{Stereo Rectification}
	3D scene geometry reconstruction with a pair of synchronized cameras is based on determining pairs of correspondence pixels between the left and right images. For an uncalibrated stereo rig, finding the correspondence pairs is a 2D search process (optical flow estimation), which is extremely computationally intensive. If the stereo rig is calibrated, 1D search should be performed along the epipolar lines. 
	An image transformation process, referred to as \textit{stereo rectification}, is always performed beforehand to reduce the dimension of the correspondence pair search. The stereo rectification consists of four main steps \cite{trucco1998introductory}:
	\begin{figure}[h!]
		\centering
		\includegraphics[width=0.55\textwidth]{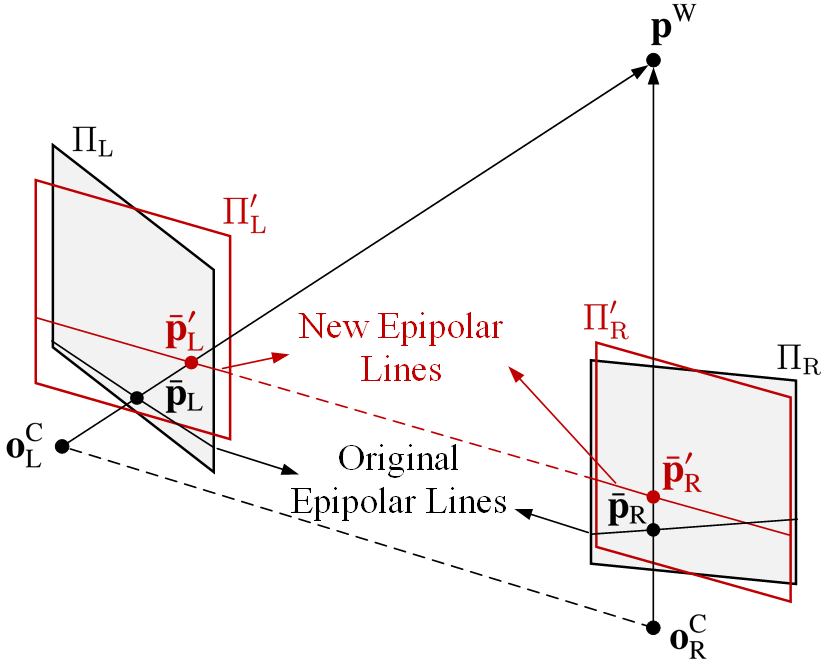}
		\caption{stereo rectification.
		}
		\vspace{-1.35em}
		\label{fig.stereo_rectification} 
	\end{figure}
	\begin{enumerate}
		\item Rotate the left camera by $\mathbf{R}_\text{rect}$ so that the left image plane is parallel to the vector $\mathbf{t}$;
		\item Apply the same rotation to the right camera to recover the original epipolar geometry;
		\item Rotate the right camera by $\mathbf{R}^{-1}$;
		\item Adjust the left and right image scales by allocating an identical intrinsic matrix to both cameras.
	\end{enumerate}
	After the stereo rectification, the left and right images appear as if they were taken by a pair of parallel cameras with the same intrinsic parameters, as shown in Fig. \ref{fig.stereo_rectification}, where $\Pi_\text{L}$ and $\Pi_\text{R}$ are the original image planes; ${\Pi'_\text{L}}$ and ${\Pi'_\text{R}}$ are the rectified image planes. Also, each pair of conjugate epipolar lines become collinear and parallel to the horizontal image axis \cite{trucco1998introductory}.  Hence, determining the correspondence pairs is simplified to a 1D search problem.

	\subsubsection{Stereo Vision System}
\begin{figure}[h!]
		\centering
		\includegraphics[width=0.61\textwidth]{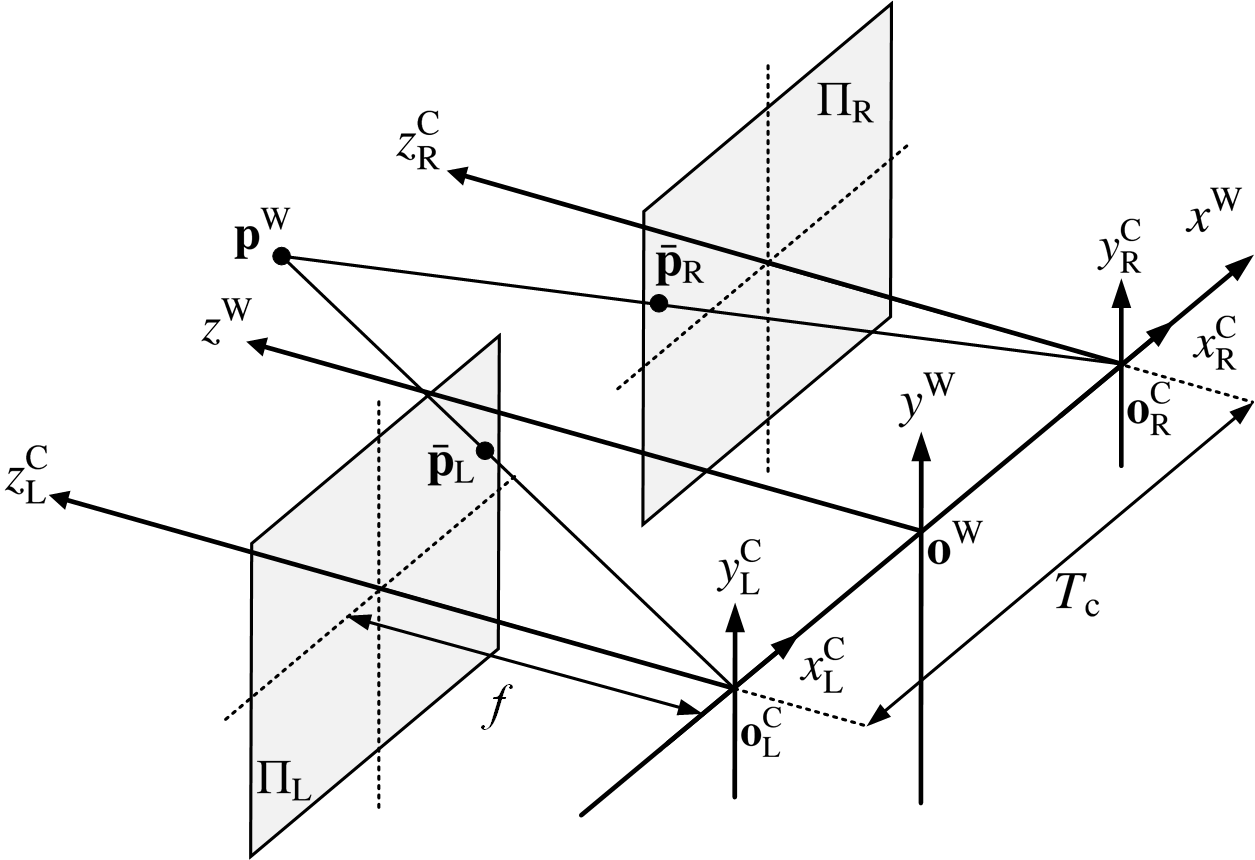}
		\caption{Basic stereo vision system.  $\mathbf{p}^\text{W}$ can be transformed to $\bar{\mathbf{p}}^\text{C}_\text{L}$ and $\bar{\mathbf{p}}^\text{C}_\text{R}$ using (\ref{eq.p_bar_pw}). }
		\label{fig.stereo_rig} 
		\vspace{-1em}
	\end{figure}

\begin{figure}[h!]
		\centering
		\includegraphics[width=0.950\textwidth]{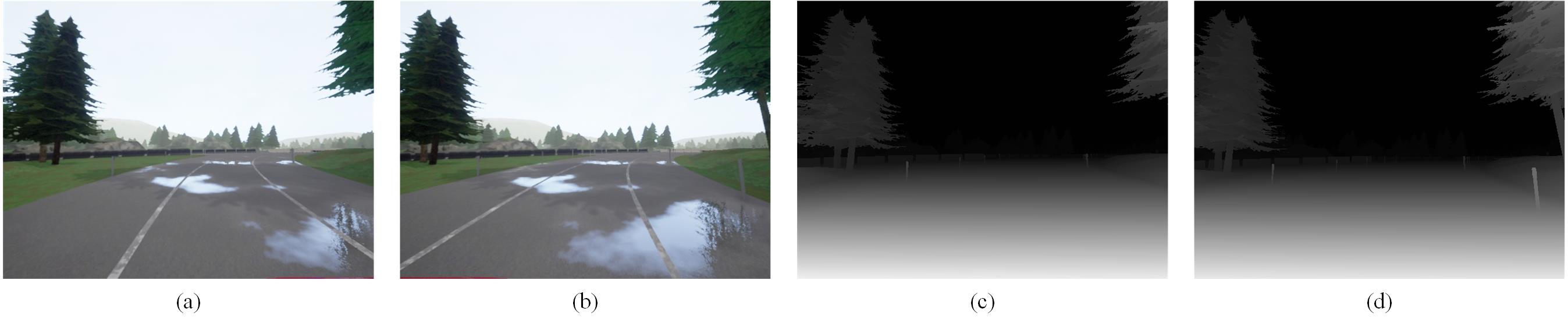}
		\caption{ (a) left image, (b) right image, (c) left disparity image $\mathbf{D}_\text{L}$ and (d) right disparity image $\mathbf{D}_\text{R}$. }
		\label{fig.disparity_estimation} 
		\vspace{-3.5em}
	\end{figure}

	A well-rectified stereo vision system is illustrated in Fig. \ref{fig.stereo_rig},
	which can be regarded as a special epipolar geometry introduced in Sec. \ref{sec.epipolar_geometry}, where the left and right cameras are perfectly parallel to each other. $x_\text{L}^\text{C}$ and $x_\text{R}^\text{C}$ axes are collinear. $\mathbf{o}_\text{L}^\text{C}$ and $\mathbf{o}_\text{L}^\text{C}$ are the left and right camera optical centers, respectively. The baseline of the stereo rig $T_c$, is defined as the distance between $\mathbf{o}_\text{L}^\text{C}$ and $\mathbf{o}_\text{R}^\text{C}$. The intrinsic matrices of the left and right cameras are given by:
	\begin{equation}
		\mathbf{K}_\text{L}=\mathbf{K}_\text{R}=\mathbf{K}=\begin{bmatrix}
			f & 0 & u_o\\
			0 & f & v_o\\
			0 & 0 & 1
		\end{bmatrix},
	\end{equation}
	respectively. Let  $\mathbf{p}^\text{W}=[x^\text{W}, y^\text{W}, z^\text{W}]^\top$ be a 3D point of interest in the WCS. Its representations in the LCCS and RCCS are  $\mathbf{p_\text{L}^\text{C}}=[x_\text{L}^\text{C}, y_\text{L}^\text{C}, z_\text{L}^\text{C}]^\top$ and $\mathbf{p_\text{R}^\text{C}}=[x_\text{R}^\text{C}, y_\text{R}^\text{C}, z_\text{R}^\text{C}]^\top$, respectively. Since the left and right cameras are considered to be exactly the same in a well-rectified stereo vision system, $s_x$ and $s_y$ in (\ref{eq.ccs2ics}) are simply set to $1$ and $f_x=f_y=f$.
	$\mathbf{p}^\text{W}$ is projected on $\Pi_\text{L}$ and $\Pi_\text{R}$ at $\bar{\mathbf{p}}_\text{L}=[x_\text{L},y_\text{L},f]^\top$ and  $\bar{\mathbf{p}}_\text{R}=[x_\text{R},y_\text{R},f]^\top$, respectively. 
	$\mathbf{o}^\text{W}$, the origin of the WCS, is at the center of the line segment $L=\{t\mathbf{{o}_\text{L}^\text{C}}+(1-t)\mathbf{{o}_\text{R}^\text{C}}\ |\ t\in[0,1]\}$. $z^\text{W}$ axis is parallel to the camera optical axes and  perpendicular to $\Pi_\text{L}$ and $\Pi_\text{R}$. Therefore, an arbitrary point $\mathbf{p^\text{W}}$ in the WCS can be transformed to $\mathbf{p}_\text{L}^\text{C}$ and $\mathbf{p}_\text{R}^\text{C}$ using: 
	\begin{equation}
		\mathbf{p}_\text{L}^\text{C}=\mathbf{I}\mathbf{p}^\text{W}+\mathbf{t}_\text{L}, \ \ \ \
		\mathbf{p}_\text{R}^\text{C}=\mathbf{I}\mathbf{p}^\text{W}+\mathbf{t}_\text{R}, 
		\label{eq.pw2pl_pr}
	\end{equation}
	where $\mathbf{t}_\text{L}=[\frac{T_c}{2},0,0]^\top$ and $\mathbf{t}_\text{R}=[-\frac{T_c}{2},0,0]^\top$; 
	Applying  (\ref{eq.normalised_image_point}) and (\ref{eq.left_right_normalised_image_point}) to (\ref{eq.pw2pl_pr}) results in the following expressions:  
	\begin{equation}
		\begin{split}
			x_\text{L}=f\frac{x^\text{W}+{T_c}/{2}}{z^\text{W}}, \ \ \ y_\text{L}=f\frac{y^\text{W}}{z^\text{W}},\\
			x_\text{R}=f\frac{x^\text{W}-{T_c}/{2}}{z^\text{W}}, \ \ \ y_\text{R}=f\frac{y^\text{W}}{z^\text{W}}.
		\end{split}
		\label{eq.p_bar_pw}
	\end{equation}
	Applying (\ref{eq.p_bar_pw}) to (\ref{eq.ccs2ics}) yields the following expressions: 
	\begin{equation}
		\mathbf{p}_\text{L}=
		\begin{bmatrix}
			u_\text{L}\\v_\text{L}
		\end{bmatrix}=
		\begin{bmatrix}
			f \frac{x^\text{W}}{z^\text{W}}+u_o +f \frac{{T_c}}{2z^\text{W}}\\
			f \frac{y^\text{W}}{z^\text{W}} +v_o
		\end{bmatrix}, \ \ \
		\mathbf{p}_\text{R}=
		\begin{bmatrix}
			u_\text{R}\\v_\text{R}
		\end{bmatrix}=
		\begin{bmatrix}
			f \frac{x^\text{W}}{z^\text{W}}+u_o -f \frac{{T_c}}{2z^\text{W}}\\
			f \frac{y^\text{W}}{z^\text{W}} +v_o
		\end{bmatrix}.
	\end{equation}
	The relationship between the so-called disparity $d$ and depth $z^\text{W}$ is as follows \cite{fan2018real}:
	\begin{equation}
		d=u_\text{L}-u_\text{R}=f\frac{T_c}{z^\text{W}}. 
		\label{eq.disparity}
	\end{equation}
	It can be observed that $d$ is inversely proportional to $z^\text{W}$. Therefore, for a distant 3D point $\mathbf{p^\text{W}}$, $\mathbf{p}_\text{L}$ and $\mathbf{p}_\text{R}$ are close to each other. On the other hand, when $\mathbf{p^\text{W}}$ lies near the stereo camera rig, the position difference between $\mathbf{p}_\text{L}$ and  $\mathbf{p}_\text{R}$ is large. Therefore, disparity estimation can be regarded as a task of 1) finding the correspondence ($\mathbf{p}_\text{L}$ and  $\mathbf{p}_\text{R}$)  pairs, which are on the same image row, on the left and right images and 2) producing two disparity images $\mathbf{D}_\text{L}$ and  $\mathbf{D}_\text{R}$, as shown in Fig. \ref{fig.disparity_estimation}. 

	\subsubsection{Disparity Estimation}
	The two key aspects of computer stereo vision are speed and accuracy \cite{tippetts2016review}. Over the past two decades, 
	a lot of research has been carried out to improve disparity estimation accuracy while reducing computational complexity. However, the stereo vision algorithms designed to achieve better disparity accuracy typically have higher computational complexity \cite{fan2018real}. Hence, speed and accuracy are two desirable but conflicting properties. It is very challenging to achieve both of them simultaneously \cite{tippetts2016review}. 
	
	In general, the main motivation of designing a stereo vision algorithm is to improve the trade-off between speed and accuracy. In most circumstances, a desirable trade-off entirely depends on the target application \cite{tippetts2016review}. For instance, a real-time performance is required for stereo vision systems employed in autonomous driving, because other systems, such as data-fusion semantic driving scene segmentation, usually take up only a small portion of the processing time, and can be easily implemented in real-time if the 3D information is available \cite{fan2018real}. Although stereo vision execution time can definitely be reduced with future HW advances, algorithm and SW improvements are also very important \cite{tippetts2016review}. 
	
	State-of-the-art stereo vision algorithms can be classified as either computer vision-based or machine/deep learning-based. The former typically formulates disparity estimation as a local block matching problem or a global energy minimization problem \cite{fan2018road}, while the latter basically considers disparity estimation as a regression problem \cite{luo2016efficient}. 
	
	\paragraph{\textbf{Computer vision-based stereo vision algorithms}}
	
	Computer vision-based disparity estimation algorithms are categorized as: 1) local, 2) global and 3) semi-global \cite{scharstein2003high}. Local algorithms simply select an image block from the left image and match it with a series of image blocks selected from the right image. Optimal disparity estimation corresponds to either the lowest difference costs or the highest correlation costs. 
	Global algorithms translate disparity estimation
	into a probability maximization problem or an energy minimization problem \cite{mozerov2015accurate}, which can be solved using Markov random field (MRF)-based optimization algorithms \cite{tappen2003comparison}. Semi-global matching (SGM) \cite{hirschmuller2007stereo} approximates MRF inferences by performing cost aggregation along all image directions, which greatly improves both disparity estimation accuracy and efficiency. Generally, a computer vision-based disparity estimation algorithm consists of four main steps: 1) cost computation, 2) cost aggregation, 3) disparity optimization and 4) disparity refinement \cite{vzbontar2016stereo}.
	\subparagraph{{1. Cost computation}}
	Disparity $d$ is a random variable with $N$ possible discrete states, each of them being associated with a matching cost $c$. The two most commonly used pixel-wise matching costs are the absolute difference (AD) cost $c_\text{AD}$ and the squared difference (SD) cost $c_\text{SD}$ \cite{vzbontar2016stereo}. Since the left and right images are typically in gray-scale format, $c_\text{AD}$ and $c_\text{SD}$ can be computed using \cite{hirschmuller2008evaluation}:
	\begin{equation}
		\begin{split}
			c_\text{AD}(\mathbf{p},d)&=\big|i_\text{L}(\mathbf{p})-i_\text{R}(\mathbf{p}-\mathbf{d})\big|, \\
			c_\text{SD}(\mathbf{p},d)&=\big(i_\text{L}(\mathbf{p})-i_\text{R}(\mathbf{p}-\mathbf{d})\big)^2, 
		\end{split}
		\label{eq.AD_SD}
	\end{equation}
	where $\mathbf{d}=[d,0]^\top$, $i_\text{L}(\mathbf{p})$ denotes the pixel intensity of $\mathbf{p}=[u,v]^\top$ in the left image and $i_\text{R}(\mathbf{p}-\mathbf{d})$ represents the pixel intensity of $\mathbf{p}-\mathbf{d}=[u-d,v]^\top$ in the right image. 
	\subparagraph{{2. Cost aggregation}}
	In order to minimize incorrect matches, pixel-wise difference costs are often aggregated over all pixels within a support region \cite{scharstein2003high}: 
	\begin{equation}
		c_\text{agg}(\mathbf{p},d)=w(\mathbf{p},d)*C(\mathbf{p},d),
		\label{eq.cost_aggregation}
	\end{equation}
	where the center of the support region is at  $\mathbf{p}=[u,v]^\top$. The corresponding disparity is $d$. $c_\text{agg}$ denotes the aggregated cost.  $w$ is a kernel that represents the support region. $C$ represents a neighborhood system containing the pixel-wise matching costs of all pixels within the support region. $c_\text{agg}$ can be obtained by performing a convolution between $w$ and $C$. A large support region can help reduce disparity optimization uncertainties, but also increase the algorithm execution time significantly. 
	
	Since the support regions are always rectangular blocks, these algorithms are also known as \textit{stereo block matching} \cite{fan2018road}. When the convolution process is a uniform box filtering (all the elements in $w$ are 1), the aggregations of $c_\text{AD}$ and $c_\text{SD}$ are referred to as the sum of absolute difference ({SAD}) and the sum of squared difference ({SSD}), respectively, which can be written as \cite{fan2018real}:
	\begin{equation}
		\begin{split}
			c_\text{SAD}(\mathbf{p},d)&={\sum\limits_{\mathbf{q}\in\mathscr{N}_\mathbf{p}}\big|i_\text{L}(\mathbf{q})-i_\text{R}(\mathbf{q}-\mathbf{d}))\big|},\\
			c_\text{SSD}(\mathbf{p},d)&={\sum\limits_{\mathbf{q}\in\mathscr{N}_\mathbf{p}}\big(i_\text{L}(\mathbf{q})-i_\text{R}(\mathbf{q}-\mathbf{d})\big)^2}, 
			\label{eq.SSD}
		\end{split}
	\end{equation}
	where $\mathscr{N}_\mathbf{p}$ is the support region (or neighborhood system) of $\mathbf{p}$. 
	Although the SAD and the SSD are computationally efficient, they are very sensitive to image intensity noise. In this regard, some other cost or similarity functions, such as the normalized cross-correlation (NCC), are more prevalently used for cost computation and aggregation. The cost function of the NCC is as follows \cite{fan2018road}:
	\begin{equation}
		c_\text{NCC}(\mathbf{p},d)=\frac{1}{n\sigma_\text{L}\sigma_\text{R}}{\sum\limits_{\mathbf{q}\in\mathscr{N}_\mathbf{p}}
			\Big(i_\text{L}\big(\mathbf{q}\big)-{\mu_\text{L}}\Big)\Big(i_\text{R}\big(\mathbf{q}-\mathbf{d}\big)-{\mu_\text{R}}\Big)},
		\label{eq.NCC}
	\end{equation}
	where
	\begin{equation}
		\sigma_\text{L}=\sqrt{\sum\limits_{\mathbf{q}\in\mathscr{N}_\mathbf{p}}\Big(i_\text{L}(\mathbf{q})-{\mu_\text{L}}\Big)^2/n}, \ \ \ \ \sigma_\text{R}=\sqrt{\sum\limits_{\mathbf{q}\in\mathscr{N}_\mathbf{p}}\Big(i_\text{R}(\mathbf{q}-\mathbf{d})-{\mu_\text{R}}\Big)^2/n},
		\label{eq.sigma_l}
	\end{equation}
	$\mu_\text{L}$ and $\mu_\text{R}$ represent the means of the pixel intensities within the left and right image block, respectively. $\sigma_\text{L}$ and $\sigma_\text{R}$ denote the standard deviations of the left and right image block, respectively. $n$ represents the number of pixels within each image blocks. 
	The NCC cost $c_\text{NCC}\in[-1,1]$ reflects the similarity between the given pair of left and right image blocks. A higher $c_\text{NCC}$ corresponds to a better block matching.
	
	In addition to the cost aggregation via uniform box filtering, many adaptive cost aggregation strategies have been proposed to improve disparity accuracy. One of the most famous algorithms is fast bilateral stereo (FBS) \cite{yang2008stereo, fan2018fbs}, which uses a bilateral filter to aggregate the matching costs adaptively. A general expression of cost aggregation in FBS is as follows \cite{fan2018ist}:
	\begin{equation}
		c_\text{agg}(\mathbf{p},d)=\frac{\sum_{\mathbf{q}\in\mathscr{N}_\mathbf{q}} \omega_d(\mathbf{q})\omega_r(\mathbf{q})c(\mathbf{q},d)}{\sum_{\mathbf{q}\in\mathscr{N}_\mathbf{q}} \omega_d(\mathbf{q})\omega_r(\mathbf{q})},
		\label{eq.fbs}
	\end{equation}
	where functions $\omega_d$ and $\omega_r$ are based on spatial distance and color similarity, respectively \cite{fan2018fbs}. The costs $c$ within a rectangular block are aggregated adaptively to produce $c_\text{agg}$.

	\subparagraph{{3. Disparity Optimization}}
	The local algorithms 
	simply select the disparities that correspond to the lowest difference costs or the highest correlation costs as the best disparities in a Winner-Take-All (WTA) way. 
	
\begin{figure}[h!]
		\centering
		\includegraphics[width=0.410\textwidth]{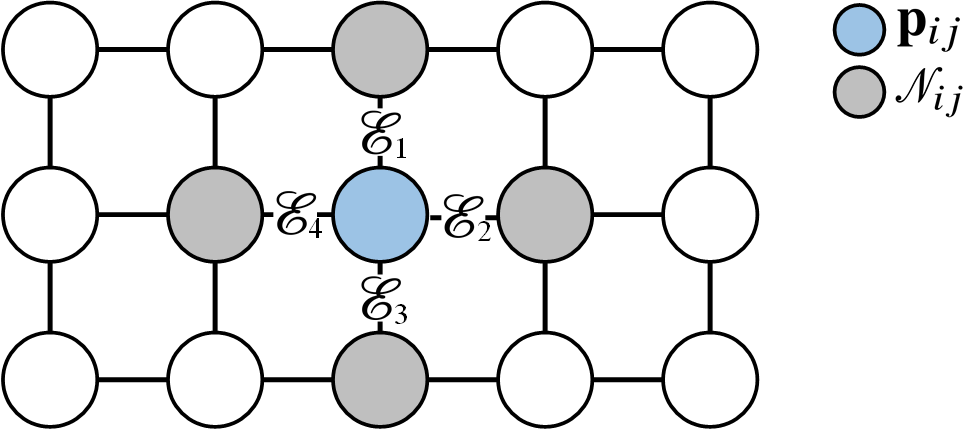}
		\caption{MRF model. }
		\label{fig.mrf} 
		\vspace{-1.2em}
	\end{figure}

	Unlike WTA applied in the local algorithms, matching costs from neighboring pixels are also taken into account in the global algorithms, \textit{e.g.}, graph cuts (GC) \cite{boykov2001fast} and belief propagation (BP) \cite{ihler2005loopy}. The MRF is a commonly used graphical model in such algorithms. An example of the MRF model is depicted in Fig. \ref{fig.mrf}.
	The graph  $\mathscr{G}=(\mathscr{P},\mathscr{E})$ is a set of vertices $\mathscr{P}$ connected by edges $\mathscr{E}$, where $\mathscr{P}=\{\mathbf{p}_{11},\mathbf{p}_{12},\cdots,\mathbf{p}_{mn}\}$ and $\mathscr{E}=\{(\mathbf{p}_{ij},\mathbf{p}_{st})\ |\ \mathbf{p}_{ij},\mathbf{p}_{st}\in\mathscr{P}\}$.  
	Two edges sharing one common vertex are called a pair of adjacent edges \cite{blake2011markov}. Since the MRF is considered to be undirected,  $(\mathbf{p}_{ij},\mathbf{p}_{st})$ and $(\mathbf{p}_{st},\mathbf{p}_{ij})$ refer to the same edge here. $\mathscr{N}_{ij}=\{\mathbf{q}_{{1}_{\mathbf{p}_{ij}}},\mathbf{q}_{{2}_{\mathbf{p}_{ij}}},\cdots,\mathbf{q}_{{k}_{\mathbf{p}_{ij}}}\ |\ \mathbf{q}_{{\mathbf{p}_{ij}}}\in\mathscr{P}\}$ is a neighborhood system of $\mathbf{p}_{ij}$.
	
	For stereo vision problems, $\mathscr{P}$ is a $m\times n$ pixel disparity image and $\mathbf{p}_{ij}$ is a graph vertex (or node) at the site of $(i,j)$ with a disparity node value $d_{ij}$.  Because the consideration of more candidates usually makes true disparity inference intractable, only the neighbors adjacent to $\mathbf{p}_{ij}$ are considered for stereo matching \cite{tappen2003comparison}, in a pairwise MRF fashion, as the disparity of node $\mathbf{p}_{ij}$ tends to have a strong correlation with its vicinities, while it is linked implicitly to any other random nodes in the disparity map. The joint MRF probability can be written as \cite{tappen2003comparison}:
	\begin{equation}
		P(\mathbf{p}, q)=\prod_{\mathbf{p}_{ij}\in\mathscr{P}} \Phi(\mathbf{p}_{ij}, q_{\mathbf{p}_{ij}})  
		\prod_{\mathbf{q}_{\mathbf{p}_{ij}}\in\mathscr{N}_{ij}} \Psi (\mathbf{p}_{ij}, \mathbf{q}_{\mathbf{p}_{ij}}),
		\label{eq.mrf_eq2}   
	\end{equation}
	where $q_{\mathbf{p}_{ij}}$ represents image intensity differences, $\Phi(\cdot)$ expresses the compatibility between  possible disparities and the corresponding image intensity differences, while $\Psi(\cdot)$ expresses the compatibility between $\mathbf{p}_{ij}$ and its neighborhood system. Now, the aim of finding the best disparity is equivalent to maximizing $P(\mathbf{p}, q)$ in (\ref{eq.mrf_eq2}), by formulating it as an energy function \cite{fan2019cvprw}: 
	\begin{equation}
		\begin{split}
			E(\mathbf{p})=\sum_{\mathbf{p}_{ij}\in\mathscr{P}} D(\mathbf{p}_{ij}, q_{\mathbf{p}_{ij}})+
			\sum_{\mathbf{q}_{\mathbf{p}_{ij}}\in\mathscr{N}_{ij}} V (\mathbf{p}_{ij}, \mathbf{q}_{\mathbf{p}_{ij}}),
		\end{split}
		\label{eq.mrf_eq3}   
	\end{equation}
	where $D(\cdot)$ and $V(\cdot)$ are two energy functions. $D(\cdot)$ corresponds to the matching cost and $V(\cdot)$ determines the aggregation from the neighbors. In the MRF model, the method to formulate an adaptive $V(\cdot)$ is important, because image intensity in discontinuous areas usually varies greatly from that of its neighbors \cite{fan2018real}. However, the process of minimizing   (\ref{eq.mrf_eq3}) results in high computational complexities, rendering real-time performance challenging. Therefore, SGM \cite{hirschmuller2007stereo} breaks down (\ref{eq.mrf_eq3}) into:
	\begin{equation}
		\begin{split}
			E(\mathbf{D})=\sum_{\mathbf{p}}\bigg(c(\mathbf{p},d_{\mathbf{p}})+\sum_{\mathbf{q}\in\mathscr{N}_\mathbf{p}}\lambda_1\delta(|d_{\mathbf{p}}-d_{\mathbf{q}}|=1)
			+\sum_{\mathbf{q}\in\mathscr{N}_\mathbf{p}}\lambda_2\delta(|d_{\mathbf{p}}-d_{\mathbf{q}}|>1)\bigg),
		\end{split}
		\label{eq.sgm_energy}
	\end{equation}
	where $\mathbf{D}$ is the disparity image, $c$ is the matching cost,  $\mathbf{q}$ is a pixel in the neighborhood system $\mathscr{N}_\mathbf{p}$ of $\mathbf{p}$. $\lambda_1$ penalizes the neighboring pixels with small disparity differences, \textit{i.e.}, one pixel; $\lambda_2$ penalizes the neighboring pixels with large disparity differences, \textit{i.e.}, larger than one pixel. $\delta(\cdot)$ returns 1 if its argument is true and 0 otherwise.
	
	\subparagraph{{4. Disparity Refinement}}
	
	Disparity refinement usually involves several post-processing steps, such as the left-and-right disparity  consistency check (LRDCC), subpixel enhancement and weighted median filtering \cite{scharstein2002taxonomy}. The LRDCC can remove most of the occluded areas, which are only visible in one of the left/right image \cite{fan2018road}. In addition, a disparity error larger than one pixel may result in a non-negligible 3D geometry reconstruction error \cite{fan2018road}. Therefore, subpixel enhancement provides an easy way to increase disparity image resolution by simply interpolating the matching costs around the initial disparity \cite{scharstein2002taxonomy}. Moreover, a median filter can be applied to the disparity image to fill the holes and remove the incorrect matches \cite{scharstein2002taxonomy}. However, the above disparity refinement algorithms are not always necessary and the sequential use of these steps depends entirely on the chosen algorithm and application needs. 
	
	\paragraph{\textbf{Machine/deep learning-based stereo vision algorithms}}
	
\begin{figure}[h!]
		\centering
		\includegraphics[width=0.70\textwidth]{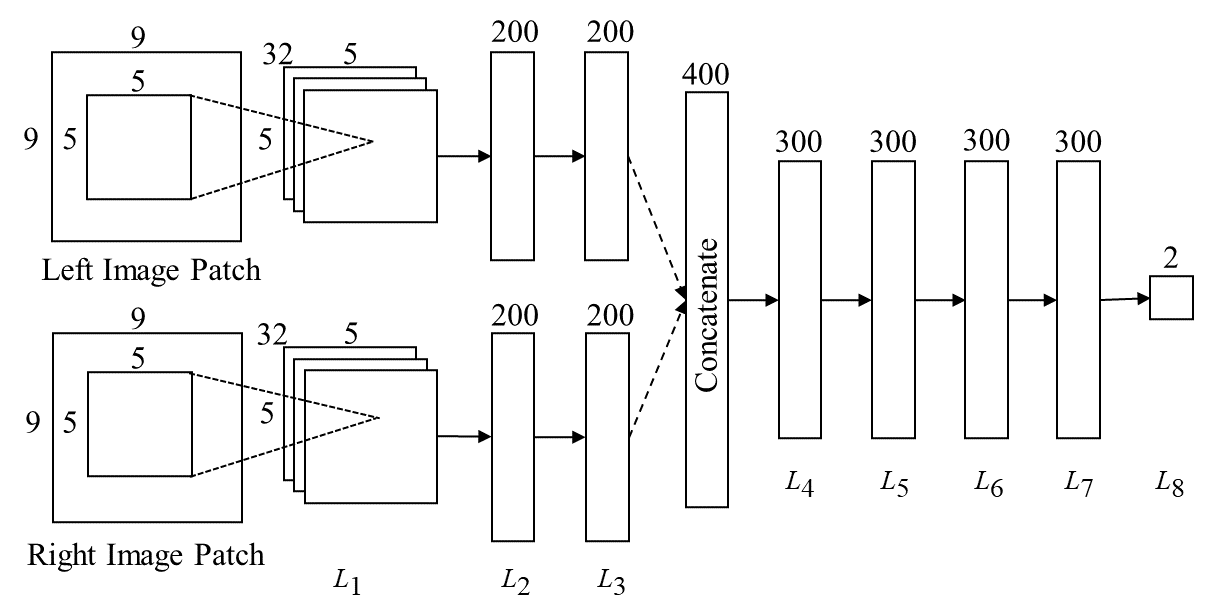}
		\caption{The architecture of the CNN proposed in \cite{zbontar2015computing} for stereo matching. }
		\label{fig.dispnet} 
		\vspace{-1.5em}
	\end{figure}

	With recent advances in machine/deep learning, CNNs have been prevalently used for disparity estimation. For instance, {\v{Z}}bontar and LeCun \cite{zbontar2015computing} utilized a CNN to compute patch-wise similarity scores, as shown in Fig. \ref{fig.dispnet}. It consists of a convolutional layer $L_1$ and seven fully-connected layers $L_2$-$L_8$. The inputs to this CNN are two 9$\times$9-pixel gray-scale
	image patches. $L_1$ consists of 32 convolution kernels of size 5$\times$5$\times$1. $L_2$ and $L_3$ have 200 neurons each. After $L_3$, the two 200-dimensional feature vectors are concatenated into a 400-dimensional vector and passed through $L_4$-$L_7$ layers. Layer $L_8$ maps $L_7$ output into two real numbers, which are then fed through a softmax function to produce a distribution over the two classes: a) good match and b) bad match. Finally, they utilize computer vision-based cost aggregation and disparity optimization/refinement techniques to produce the final disparity images. Although this method has achieved the  state-of-the-art accuracy, it is limited by the employed matching cost aggregation technique and can produce wrong predictions in occluded or texture-less/reflective regions \cite{zhang2019GANet}.

\begin{figure}[h!]
		\vspace{-1.0em}
		\centering
		\includegraphics[width=0.61\textwidth]{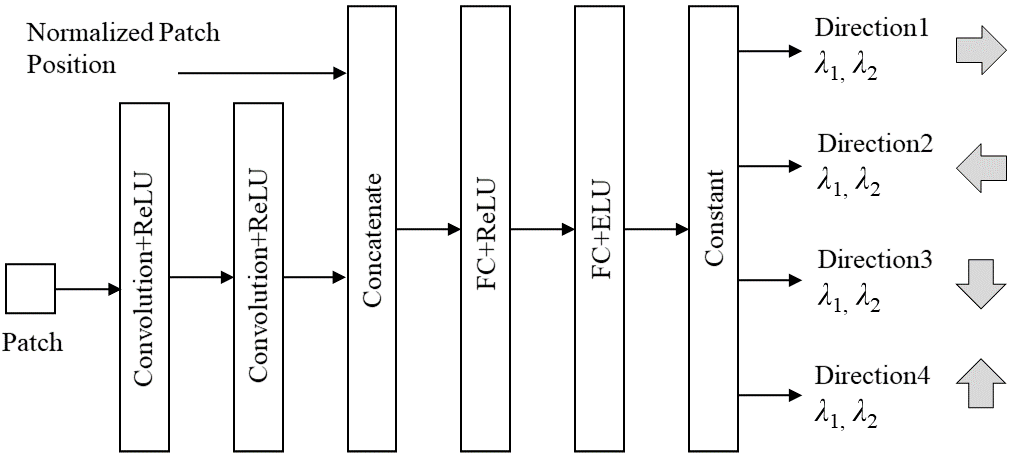}
		\caption{SGM-Nets \cite{seki2017sgm} architecture. }
		\label{fig.sgmnets} 
		\vspace{-1.5em}
	\end{figure}

	In this regard, some researchers have leveraged CNNs to improve computer vision-based cost aggregation step. SGM-Nets \cite{seki2017sgm} is one of the most well-known methods of this type. Its main contribution is a CNN-based technique for predicting SGM penalty parameters $\lambda_1$ and $\lambda_2$ in (\ref{eq.sgm_energy}) \cite{hirschmuller2007stereo}, as illustrated in Fig. \ref{fig.sgmnets}. A $5\times5$-pixel gray-scale image patch and its normalized position are used as the CNN inputs. It has a) two convolution layers, each followed by a rectified linear unit (ReLU) layer; b) a concatenate layer for merging the two types of inputs; c) two fully connected (FC) layers of size 128 each, followed by a ReLU layer and an exponential
	linear unit (ELU); d) a constant layer to keep SGM penalty values positive. The costs can then be  accumulated along four directions. The CNN output values correspond to standard parameterization. 
	
	Recently, end-to-end deep CNNs have
	become very popular. For example, Mayer \textit{et al.} \cite{mayer2016large} created three large synthetic datasets\footnote{\url{lmb.informatik.uni-freiburg.de/resources/datasets/SceneFlowDatasets.en.html}} ( 
	FlyingThings3D, Driving and Monkaa) and proposed a CNN named DispNet for dense disparity estimation. Later on, Pang \textit{et al.} \cite{pang2017cascade} proposed a two-stage cascade CNN for disparity estimation. Its the first stage enhances DispNet \cite{mayer2016large} by equipping it with extra up-convolution modules and the second stage rectifies the disparity initialized by the first stage and generates residual signals across multiple scales. Furthermore, GCNet \cite{kendall2017end} incorporated feature extraction (cost computation), cost aggregation and disparity optimization/refinement into a single end-to-end CNN model, and it achieved the state-of-the-art accuracy on the FlyingThings3D benchmark \cite{mayer2016large} as well as the KITTI stereo 2012 and 2015 benchmarks \cite{geiger2012we, menze2015joint, menze2018object}. In 2018, Chang \textit{et al.} \cite{chang2018pyramid} proposed Pyramid Stereo Matching Network (PSMNet),  consisting of two modules: a) spatial pyramid pooling and b) 3D CNN. The former aggregates the context of different scales and locations, while the latter regularizes the cost volume. Unlike PSMNet \cite{chang2018pyramid}, guided aggregation net (GANet) \cite{zhang2019GANet} replaces the widely used 3D CNN with two novel layers: a semi-global aggregation layer and a local guided aggregation layer, which help save a lot of memory and computational cost.

	Although the aforementioned CNN-based disparity estimation methods have achieved compelling results, they usually have a huge number of learnable parameters, resulting in a long processing time. Therefore, current state-of-the-art CNN-based disparity estimation algorithms have hardly been put into practical uses in autonomous driving. We believe these methods will be applied in more real-world applications, with future advances in embedded computing HW.
	
	\subsubsection{Performance Evaluation}
	\label{sec.algorithm_evaluation}
	As discussed above, disparity estimation speed and accuracy are two key properties and they are always pitted against each other. Therefore, the performance evaluation of a given stereo vision algorithm  usually involves both of these two properties \cite{tippetts2016review}. 
	
	The following two metrics are commonly used to evaluate the accuracy of an estimated disparity image \cite{barron1994performance}: 
	\begin{enumerate}
		\item Root mean squared (RMS) error $e_\text{RMS}$: 
		\begin{equation}
			e_\text{RMS}=\sqrt{\frac{1}{N}\sum_{\mathbf{p}\in\mathscr{P}}|\mathbf{D}_\text{E}(\mathbf{p})-\mathbf{D}_\text{G}(\mathbf{p})|^2},
			\label{eq.e_rms}
		\end{equation}
		\item Percentage of error pixels (PEP) $e_\text{PEP}$ (tolerance: $\delta_d$ pixels):
		\begin{equation}
			e_\text{PEP}={\frac{1}{N}\sum_{\mathbf{p}\in\mathscr{P}}\delta\bigg(|\mathbf{D}_\text{E}(\mathbf{p})-\mathbf{D}_\text{G}(\mathbf{p})|>\delta_d\bigg)}\times100\%,
			\label{eq.e_pep}
		\end{equation}
	\end{enumerate}
	where $\mathbf{D}_\text{E}$ and $\mathbf{D}_\text{G}$ represent the estimated and ground truth disparity images, respectively; $N$ denotes the total number of disparities used for evaluation; $\delta_d$ represents the disparity evaluation tolerance. 
	
	Additionally, a general way to depict the efficiency of an algorithm is given in millions of disparity evaluations per second $\text{Mde}/s$ \cite{tippetts2016review} as follows: 
	\begin{equation}
		\text{Mde}/s=\frac{u_\text{max}v_\text{max}d_\text{max}}{t}{10^{-6}}.
	\end{equation}
	However, the speed of a disparity estimation algorithm typically varies across different platforms, and it can be greatly boosted by exploiting the parallel computing architecture. 
	
	\section{Heterogeneous Computing}
	\label{sec.heterogeneous_system}

\begin{figure}[h!]
		\centering
		\includegraphics[width=0.58\textwidth]{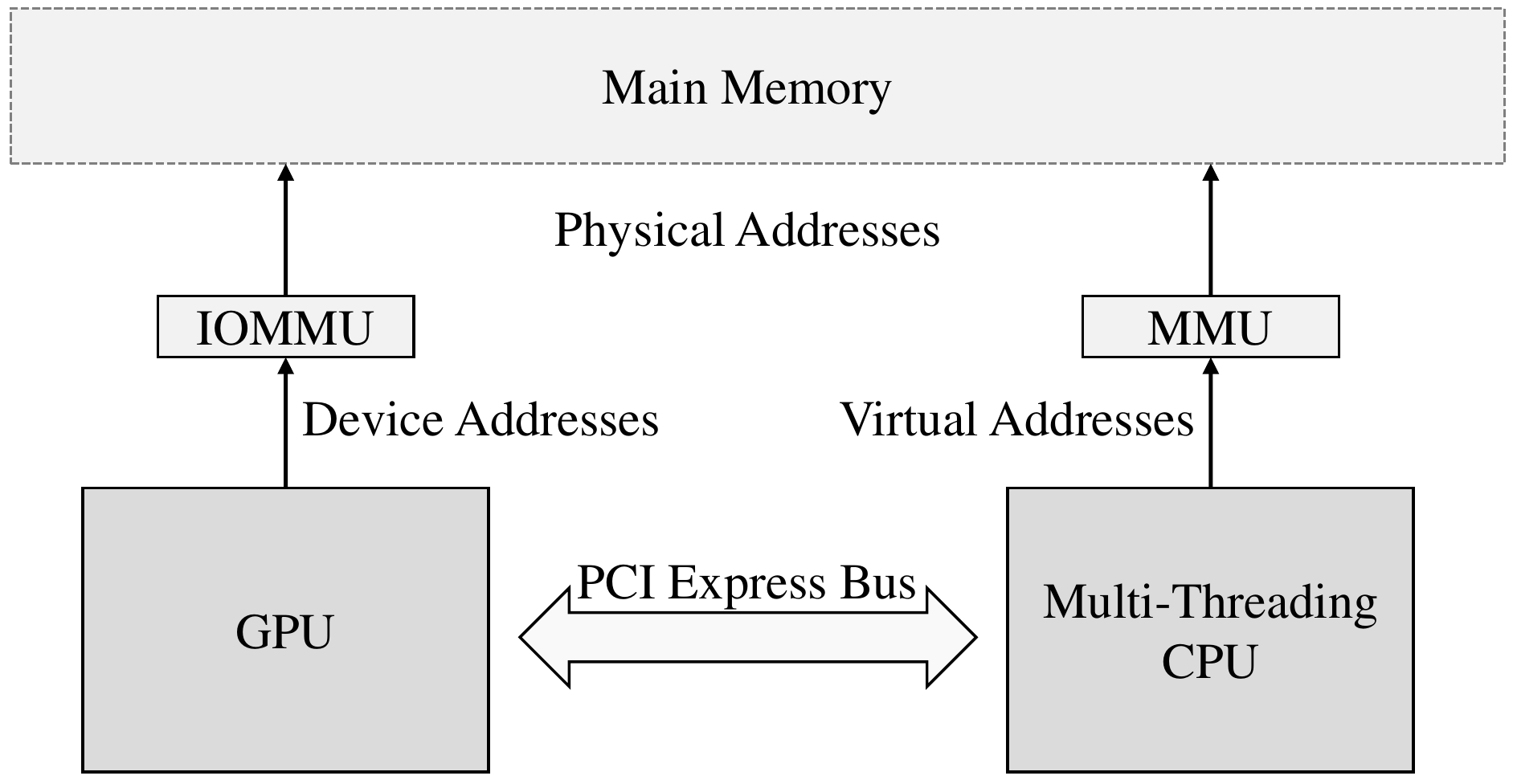}
		\caption{Heterogeneous system architecture.}
		\label{fig.heterogeneous_system} 
		\vspace{-1.5em}
	\end{figure}

	Heterogeneous computing systems use multiple types of processors or cores. In the past, heterogeneous computing meant that different instruction-set architectures (ISAs) had to be handled differently, while  modern heterogeneous system architecture (HSA) systems allow users to utilize multiple processor types. As illustrated in Fig. \ref{fig.heterogeneous_system}, 
	a typical HSA system consists of two different types of processors: 1) a multi-threading central processing unit (CPU) and 2) a graphics processing unit (GPU) \cite{mittal2015survey}, which are connected by a peripheral component interconnect (PCI) express bus. The CPU's memory management unit (MMU) and the GPU's input/output memory management unit (IOMMU) comply with the HSA HW specifications. CPU runs the operating system and performs traditional serial computing tasks, while GPU performs 3D graphics rendering and CNN training.

	\subsection{Multi-Threading CPU}
\begin{figure}[h!]
		\centering
		\includegraphics[width=0.72\textwidth]{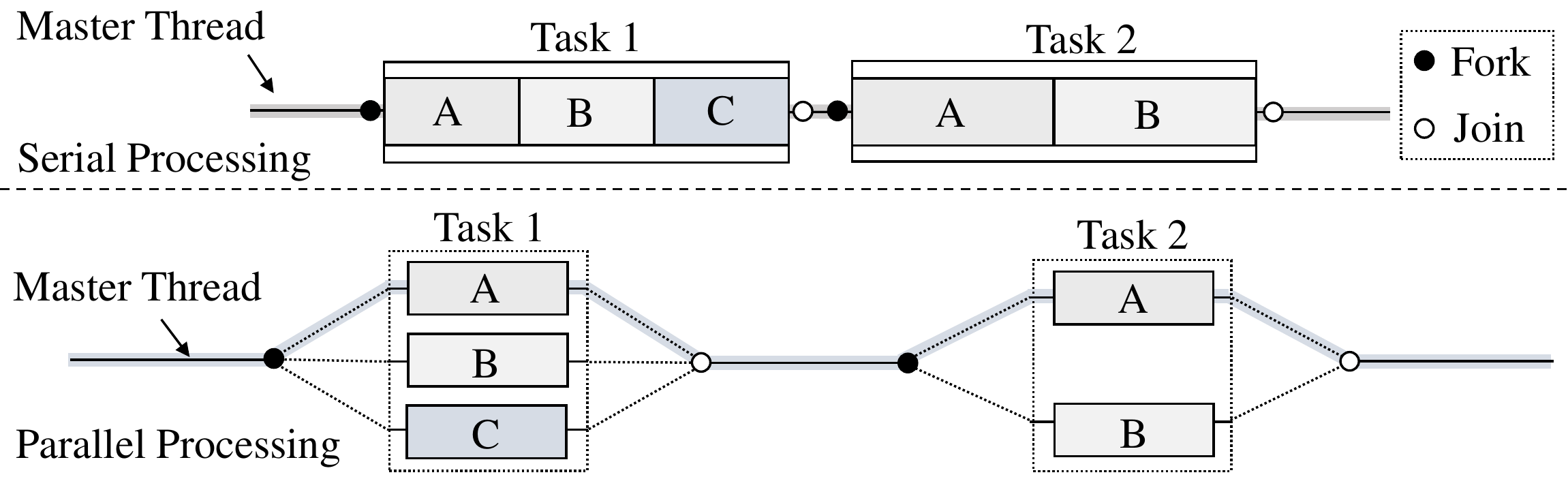}
		\caption{Serial processing vs. parallel processing.}
		\label{fig.openmp} 
		\vspace{-1.5em}
	\end{figure}

	The application programming interface (API) Open Multi-Processing (OpenMP) is typically used to break a serial code into independent chunks for parallel processing. It supports multi-platform shared-memory multiprocessing programming in C/C++ and Fortran \cite{jin2011high}. An explicit parallelism programming model, typically known as a fork-join model,  is illustrated in Fig. \ref{fig.openmp}, 
	where the compiler instructs a section of the serial code to run in parallel \cite{fan2019icvs}. The master thread (serial execution on one core) forks a number of slave threads. The tasks are divided to run in parallel amongst the slave threads on multiple cores. Synchronization  waits until all slave threads finish their allocated tasks \cite{fan2016ist}. Finally, the slave threads join together at a subsequent point and resume sequential execution.

	\subsection{GPU}
	GPUs have been extensively used in computer vision and deep learning to accelerate the computationally intensive but parallelly-efficient processing and CNN training. Compared with a {CPU}, which consists of a low number of cores optimized for sequentially serial processing, {GPU} has a highly parallel architecture which is composed of hundreds or thousands of light GPU cores to handle multiple tasks concurrently. 
	
\begin{figure}[h!]
		\centering
		\includegraphics[width=0.55\textwidth]{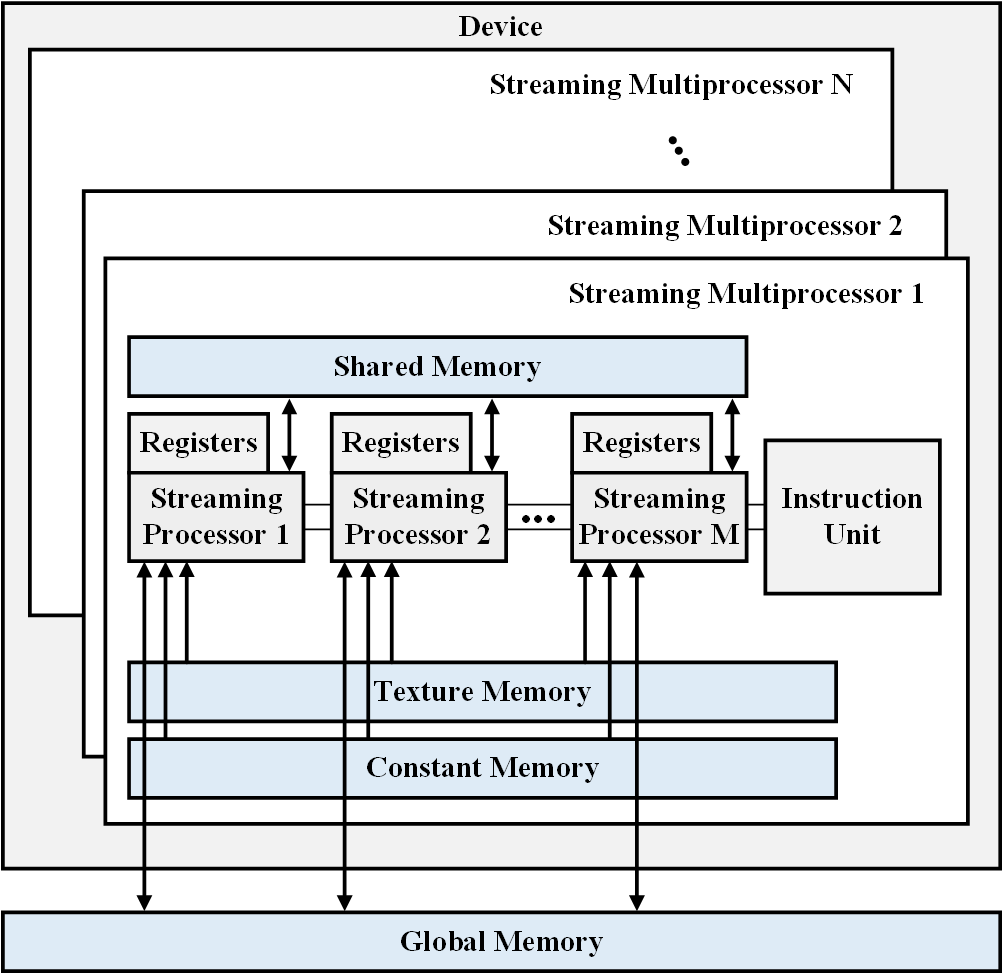}
		\caption{GPU architecture \cite{fan2018fbs}.}
		\label{fig.gpu} 
		\vspace{-1.5em}
	\end{figure}

	A typical GPU architecture  is shown in Fig. \ref{fig.gpu}, which consists of $N$ streaming multiprocessors (SMs) with $M$ streaming processors (SPs) on each of them. The single instruction multiple data (SIMD) architecture allows the SPs on the same SM to execute the same instruction but process different data at each clock cycle. 
	The device has its own dynamic random access memory ({DRAM}) which consists of global memory, constant memory and texture memory. DRAM can communicate with the host memory via the graphical/memory controller hub ({GMCH}) and the I/O controller hub ({ICH}), which are also known as the Intel northbridge and the Intel southbridge, respectively. Each SM has four types of on-chip memories: register, shared memory, constant cache and texture cache. Since they are on-chip memories, the constant cache and texture cache are utilized to speed up data fetching from the constant memory and texture memory, respectively. Due to the fact that the shared memory is small, it is used for the duration of processing a block. The register is only visible to the thread. The details of different types of GPU memories are illustrated in Table \ref{tab.gpu_memory}.
	
	\begin{table}[!t]
		\begin{center}	
			\footnotesize
			\caption{GPU memory comparison \cite{fan2017ist}.}
			\begin{tabular}{l|c|c|c|c}
				\hline
				Memory & Location & Cached & Access & Scope\\
				\hline
				register & on-chip & n/a & r/w & one thread \\
				shared  & on-chip & n/a & r/w & all threads in a block \\
				global & off-chip & no & r/w & all threads + host\\
				constant  & off-chip & yes & r & all threads + host\\
				texture & off-chip & yes & r & all threads + host\\
				\hline
			\end{tabular}
		\end{center}
		\vspace{-2.5em}
		\label{tab.gpu_memory}
	\end{table}

	In CUDA C programming, the threads are grouped into a set of 3D thread blocks which are then organized as a 3D  grid. The kernels are defined on the host using the CUDA C programming language. Then, the host issues the commands that submit the kernels to devices for execution. Only one kernel can be executed at a given time. Once a thread block is distributed to an SM, the threads are divided into groups of 32 parallel threads which are executed by SPs. Each group of 32 parallel threads is known as a warp. Therefore, the size of a thread block is usually chosen as a multiple of 32 to ensure efficient data processing.

	\section{Summary}
	In this chapter, we first introduced the autonomous car system, from both HW aspect (car sensors and car chassis) and SW aspect (perception, localization and mapping, prediction and planning, and control). Particularly, we introduced the autonomous car perception module, which has four main functionalities:  1) visual feature detection, description and matching, 2) 3D information acquisition, 3) object detection/recognition and 4) semantic image segmentation. Later on, we provided readers with the preliminaries for the epipolar geometry and introduced computer stereo vision from theory to algorithms. Finally, heterogeneous computing architecture, consisting of a multi-threading CPU and a GPU, was introduced. 
	
	\section*{Acknowledgments}
	This chapter has received partial funding from the European Union’s Horizon 2020 research and innovation programme under grant agreement No. 871479 (AERIALCORE).

	\bibliographystyle{IEEEtran}

\end{document}